\documentclass{article}
\PassOptionsToPackage{numbers, compress}{natbib}

\usepackage[preprint]{neurips_2024}

\usepackage[utf8]{inputenc} 
\usepackage[T1]{fontenc}    
\usepackage{hyperref}       
\usepackage{url}            
\usepackage{booktabs}       
\usepackage{amsfonts}       
\usepackage{nicefrac}       
\usepackage{microtype}      
\usepackage{xcolor}         
\usepackage{graphicx}
\usepackage{amsmath}
\usepackage{overpic}
\usepackage{wrapfig}
\usepackage{color}
\usepackage{xcolor}
\usepackage{colortbl}
\usepackage{enumitem}
\definecolor{tabfirst}{rgb}{1, 0.7, 0.7} 
\definecolor{tabsecond}{rgb}{1, 0.85, 0.7} 
\definecolor{tabthird}{rgb}{1, 1, 0.7} 
\usepackage{amsmath}
\usepackage{array}
\usepackage{multirow}

\newcommand{\ie}{i.e.\@\xspace}

\definecolor{darkviolet}{rgb}{0.58, 0.2, 0.63}

\newcommand{\remove}[1]{}

\title{GSDF: 3DGS Meets SDF for Improved Neural Rendering and Reconstruction}

\author{Mulin Yu $^{1}$\thanks{Equal contribution} \quad Tao Lu$^{1}$\protect\footnotemark[1] \quad Linning Xu$^2$ \quad Lihan Jiang$^{4, 1}$ \quad Yuanbo Xiangli$^3$\thanks{Corresponding author}
\quad Bo Dai$^{1}$\\
$^1$Shanghai Artificial Intelligence Laboratory, $^2$The Chinese University of Hong Kong, \\
$^3$Cornell University, $^4$ University of Science and Technology of China
}

\begin{document}
\newcommand{\modelname}{GSDF\xspace}

\maketitle

\begin{figure*}[!h]
  \centering
    \includegraphics[width=\linewidth]{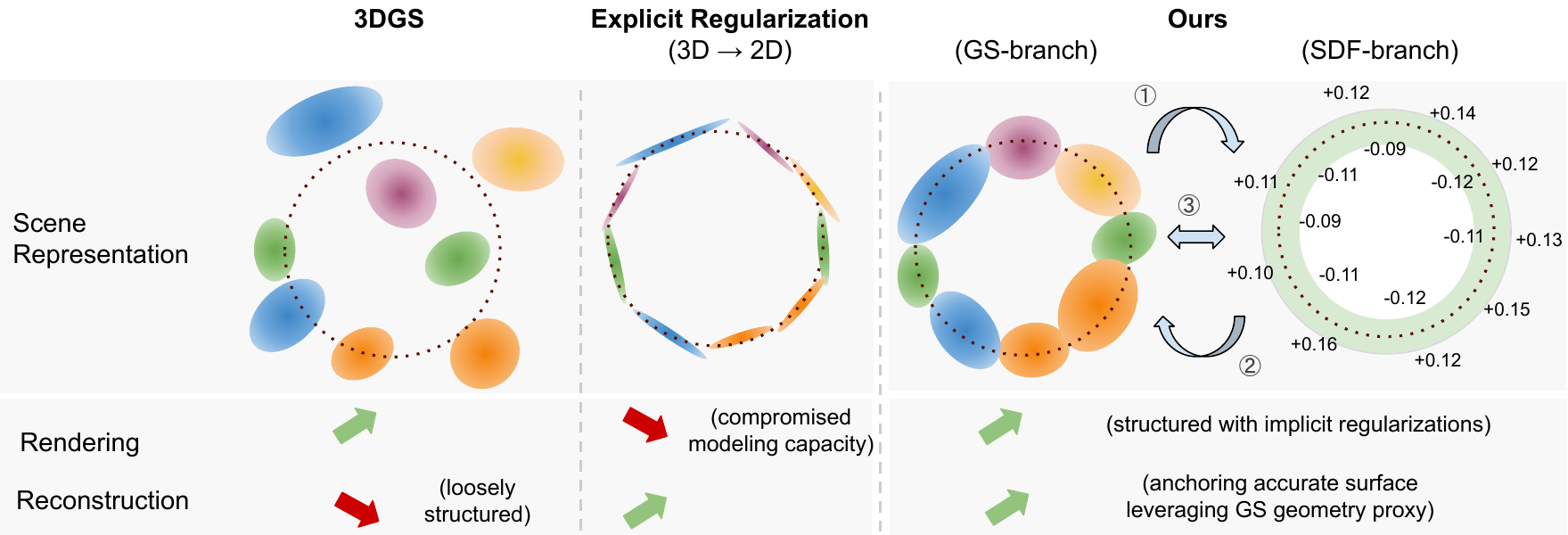}
   \caption{{\textbf{Conceptual Illustration of GSDF.} 
Rendering and reconstruction tasks have traditionally involved trade-offs in neural representation methods. While 3D-GS achieves high-fidelity view-dependent rendering, it often compromises on geometric accuracy. Recent approaches~\cite{Huang20242DGS, guedon2023sugar} use explicit regularization to align Gaussian primitives near surfaces, but this can reduce model capacity for high-fidelity visuals. Our GSDF introduces a dual-branch framework with specialized GS- and SDF-branches for rendering and geometry tasks. We propose three mutual guidances (detailed in Sec.~\ref{sec:dual_branch}) to enhance the quality of both tasks.}}
\label{fig:teaser}
\end{figure*}

\begin{abstract}

Representing 3D scenes from multiview images remains a core challenge in computer vision and graphics, requiring both reliable rendering and reconstruction, which often conflicts due to the mismatched prioritization of image quality over precise underlying scene geometry. Although both neural implicit surfaces and explicit Gaussian primitives have advanced with neural rendering techniques, current methods impose strict constraints on density fields or primitive shapes, which enhances the affinity for geometric reconstruction at the sacrifice of rendering quality. To address this dilemma, we introduce GSDF, a dual-branch architecture combining 3D Gaussian Splatting (3DGS) and neural Signed Distance Fields (SDF).
Our approach leverages mutual guidance and joint supervision \emph{during the training process} to mutually enhance reconstruction and rendering. 
Specifically, our method guides the Gaussian primitives to locate near potential surfaces and accelerates the SDF convergence. 
This implicit mutual guidance ensures robustness and accuracy in both synthetic and real-world scenarios. Experimental results demonstrate that our method boosts the SDF optimization process to reconstruct more detailed geometry, while reducing floaters and blurry edge artifacts in rendering by aligning Gaussian primitives with the underlying geometry. 

\remove{
Presenting a 3D scene from multiview images remains a core and long-standing challenge in computer vision and computer graphics.
Two main requirements lie in \emph{rendering} and \emph{reconstruction}.
Notably, SOTA rendering quality is usually achieved with neural volumetric rendering techniques, which rely on aggregated point/primitive-wise color and neglect the underlying scene geometry.
Learning of neural implicit surfaces is sparked from the success of neural rendering.
Current works either constrain the distribution of density fields or the shape of primitives,
resulting in degraded rendering quality and flaws on the learned scene surfaces.
The efficacy of such methods is limited by the inherent constraints of the chosen neural representation,
which struggles to capture fine surface details, especially for larger, more intricate scenes. 
To address these issues, we introduce \modelname, a novel dual-branch architecture that combines the benefits of a flexible and efficient 3D Gaussian Splatting (3DGS) representation with neural Signed Distance Fields (SDF).
The core idea is to leverage and enhance the strengths of each branch while alleviating their limitation through mutual guidance and joint supervision.
We show on diverse scenes that our design unlocks the potential for more accurate and detailed surface reconstructions, and at the meantime benefits 3DGS rendering with structures that are more aligned with the underlying geometry.
}

\end{abstract}

\section{Introduction}
\label{sec:intro}

Recent advancements in neural scene representations have showcased superior rendering capabilities \cite{mildenhall2020nerf, muller2022instant, kerbl20233d}, these advancements have sparked significant interest in neural surface reconstruction \cite{yariv2021volume,wang2021neus,li2023neuralangelo,Wang2022NeuS2FL,guedon2023sugar,Huang20242DGS}. They seek to develop unified representations, which simultaneously support high-fidelity rendering and accurate geometric reconstruction, to better support downstream applications such as robotics \cite{keetha2023splatam,rosinol2023nerf}, physical simulations \cite{xie2023physgaussian,feng2024gaussian}, and XR applications \cite{xu2023vr,jiang2024vr}.

Although both neural implicit surfaces and explicit Gaussian primitives have advanced with neural rendering techniques, current methods often suffer from a mismatched prioritization between appearance and geometry. Imposing regularization or constraints on the unified representation may boost one task, while unavoidably deteriorating the performance of the other. Recent approaches~\cite{guedon2023sugar,chen2023neusg}, have explored using flat Gaussian primitives for surface modeling. Enforcing binary opacity \cite{guedon2023sugar} and jointly learned NeuS models \cite{chen2023neusg} to regularize attributes have led to degraded rendering quality due to primitive constraints. However, works such as Adaptive Shell \cite{wang2023adaptive}, Binary Occupancy Field \cite{reiser2024binary}, and Scaffold-GS \cite{scaffoldgs} have demonstrated that incorporating geometry guidance significantly enhances rendering quality by producing well-regularized spatial structures with hybrid representations.

Building on these insights, we propose a synchronously optimized \emph{dual-branch} system, addressing rendering and reconstruction with hybrid
representations, to bypass the conflicts and further achieve mutual enhancements, as illustrated in Fig.~\ref{fig:teaser}. Our method leverages mutual guidance and joint supervision to balance rendering and reconstruction without compromising their intrinsic advantages. Specifically, our system features a 3D Gaussian Splatting (3DGS) branch for rendering and a Signed Distance Field (SDF) branch for surface reconstruction. The key innovations of our approach include:
(1) Utilizing rasterized depth from the GS-branch to guide ray sampling in the SDF-branch, enhancing volume rendering efficiency and avoiding local minima.
(2) Applying SDF-guidance for density control in 3DGS, directing the growth of 3D Gaussians in near-surface regions and pruning elsewhere.
(3) Aligning geometry properties (depth and normal) estimated from both branches.
This unified system overcomes the limitations inherent in each method due to differences in rendering techniques (rasterization vs. dense ray sampling) and scene representation (discrete primitives vs. continuous fields). Moreover, our framework is designed to accommodate future advancements in each branch.

\remove{
Specifically, our system incorporates a GS-branch for rendering and an SDF-branch for surface reconstruction.
Motivated by the strengths of both sides\ie, fast training for coarse geometry and efficient rasterization from the 3DGS, along with the continuous geometry prior from the neural SDF branch,
we propose:
1) Utilizing the rasterized depth from the fast GS-branch to guide ray sampling in the SDF-branch, 
enhancing the efficiency of volume rendering and avoiding local minima;
2) Employing the SDF-guidance for density control in 3D-GS, guiding the growth of 3D Gaussians in near-surface regions and pruning otherwise.
3) Aligning geometry properties (depth and normal) estimated from both branches.
We found this unified system overcomes the limitations of each sides originated from the differences between the rendering methods (i.e. rasterization vs. dense ray sampling), as well as the scene representation (i.e. discrete primitives vs. continuous fields).
Furthermore, our framework effortlessly accommodates future advancements for each branch.
}

Extensive experiments demonstrate that our dual-branch design allows:
1) The GS-branch to generate structured primitives closely aligned with the surface, reducing floaters and improving detail and edge quality in view synthesis.
2) Accelerated convergence in the SDF-branch, resulting in superior geometric accuracy and enhanced surface details.
Our results confirm that an integrated blend of both reconstruction and rendering is achievable, enhancing overall robustness and performance.

\section{Related work}
\label{sec:related_work}

\paragraph{Advanced Neural Rendering Techniques.}
Neural Radiance Fields (NeRFs) \cite{mildenhall2020nerf} have achieved remarkable photorealistic rendering with view-dependent effects. They use Multi-Layer Perceptrons (MLPs) to map 3D spatial locations to color and density, which are then aggregated into pixel colors through neural volumetric rendering. 
This approach excels in novel view synthesis but is slow due to the need for extensive point sampling along each ray and the global MLP architecture's scalability limitations.
Recent research \cite{muller2022instant,xu2023gridnerf,chen2022tensorf,fridovich2022plenoxels} has shifted the learning burden to locally optimized spatial features, enhancing scalability.
Alternative methods like rasterizing geometric primitives (e.g., point clouds) \cite{aliev2020neural,yifan2019differentiable} offer efficiency and flexibility but struggle with discontinuities and outliers. Augmenting points with neural features and incorporating volumetric rendering \cite{xu2022point} improves quality but adds computational overhead.
Recently, 3D Gaussian Splatting (3DGS) \cite{kerbl20233d} revolutionized neural rendering by using anisotropic 3D Gaussians as primitives, sorted by depth and rasterized onto a 2D screen with $\alpha$-blending. This method achieves high-quality, detailed results at real-time frame rates. Scaffold-GS \cite{scaffoldgs} enhanced 3DGS by introducing a hierarchical structure that aligns anchors with scene geometry, improving rendering quality and memory efficiency.
However, these methods prioritize view synthesis over accurate scene geometry, often resulting in fuzzy volumetric density fields that hinder the extraction of high-quality surfaces

\paragraph{Neural Surface Reconstruction.}

The success of neural rendering has sparked significant interest in neural surface reconstruction \cite{oechsle2021unisurf,yariv2021volume,wang2021neus,li2023neuralangelo,Wang2022NeuS2FL,yariv2023bakedsdf}. These methods typically use coordinate-based networks to encode scene geometry through occupancy fields or Signed Distance Field (SDF) values. While MLPs with volume rendering produce smooth and complete surfaces, they often lack high-fidelity details and are slow to optimize.
Recent works \cite{li2023neuralangelo, wang2023adaptive, reiser2024binary} have leveraged multi-resolution hashed feature grids from iNGP \cite{muller2022instant} to enhance representation power, achieving state-of-the-art results. Hybrid approaches combining surface and volume rendering \cite{turki2023hybridnerf,wang2023adaptive,reiser2024binary} have also emerged to maintain rendering speed and quality.
More recently, methods have explored integrating 3D Gaussian Splatting for surface learning. For example, SuGaR \cite{guedon2023sugar} aligns 3D Gaussians with potential surfaces by approximating them to 2D planar primitives with binary opacity. Similarly, 2D Gaussian Splatting \cite{Huang20242DGS} replaces 3D Gaussians with 2D Gaussians for more accurate ray-splat intersection and regularizes depth maps to enforce a more condensed Gaussian primitive distribution. Another approach \cite{Dai2024HighqualitySR} optimizes Gaussian surfels constrained by a normal prior from a pre-trained monocular estimator, enhancing surface representation.
The concurrent work NeuSG~\cite{chen2023neusg} jointly optimizes 3D Gaussian Splatting with NeuS~\cite{wang2021neus} by encouraging flat 3D Gaussians as well, with normals aligned with NeuS predictions. 3DGSR~\cite{lyu20243dgsr} also aligns the geometry from 3DGS with a SDF field.
Despite advances in neural surface reconstruction, a fidelity gap persists due to the explicit regularization of Gaussian primitives, which hampers rendering quality compared to 3DGS. Our method optimizes primitives aligned with surfaces using geometric clues, enhancing structural integrity without losing representational power. 

\section{Method}
\label{sec:method}

\begin{figure}[t]
  \centering
   \includegraphics[width=1\linewidth]
   {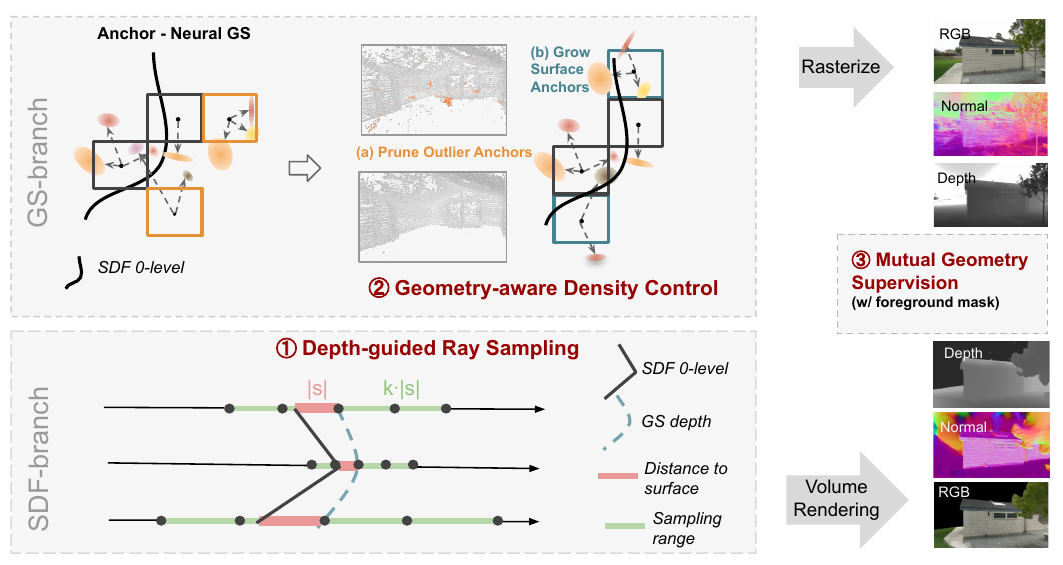}
   \caption{\textbf{Overview of Dual-branch Guidance}. 
Our dual-branch framework includes a GS-branch for rendering and an SDF-branch for learning neural surfaces. This design preserves the efficiency and fidelity of Gaussian primitive for rendering~\cite{kerbl20233d,scaffoldgs} while accurately approximating scene surfaces from an SDF field adapted from NeuS~\cite{wang2021neus}. Specifically: (1) The GS-branch renders depth maps to guide SDF-branch ray sampling, querying absolute SDF values $|s|$ and sampling points within $2k|s|$ (e.g., $k=4$).
(2) Predicted SDF values guide GS-branch density control, growing Gaussians near surfaces and pruning deviated ones.
(3) Mutual geometry consistency is enforced by comparing depth and normal maps from both branches, ensuring coherent alignment between Gaussians and surfaces.
   }

   \label{fig:pipeline}
   \vspace{-20pt}
\end{figure}

\remove{
We present a dual-branch framework featuring a GS-branch focusing on efficient and high-quality rendering, and an SDF-branch concentrates on neural implicit surface reconstruction.
In Sec.~\ref{sec:preliminary}, we first briefly go through current SOTA rendering techniques including 3D Gaussian Splatting~\cite{kerbl20233d} and Scaffold-GS~\cite{scaffoldgs}; and representative SDF-based neural implicit surface learning methods including NeuS~\cite{wang2021neus} and Neurolangelo~\cite{li2023neuralangelo}.
We show that optimizing solely for either rendering or reconstruction can lead to flaws, for example, floaters in 3DGS, and holes on the learned surface as shown in Fig.~\ref{fig:teaser}.  
We then delve into the details of our proposed dual-branch framework in Sec~\ref{sec:dual_branch}.
Particularly, we elaborate the use of depth maps rendered from the GS-branch to guide ray sampling of the SDF-branch;
and describes our geometry-aware Gaussian density control that results in better structured Gaussian primitives.
We further introduce the mutual geometry supervision, which impose an implicit and finer-grained regularization between these two branches.  
Sec.~\ref{sec:losses} provides more details on our training strategy and loss design.
}

We present a dual-branch framework with GS- and SDF-branches, jointly optimized to provide mutual guidance, as shown in Fig.~\ref{fig:pipeline}. Our approach aligns Gaussian primitives with surfaces using geometric clues rather than constraining their shapes, enhancing both structure and representational power. These improved primitives yield more accurate and detailed scene surfaces.

In Sec.\ref{sec:preliminary}, we briefly review neural rendering methods\cite{kerbl20233d,scaffoldgs} and neural implicit surface representations~\cite{wang2021neus,li2023neuralangelo}. Sec.\ref{sec:dual_branch} details our framework and the three proposed mutual guidance techniques. Sec.\ref{sec:losses} outlines our training strategy and loss design.

\subsection{Preliminary}
\label{sec:preliminary}

\paragraph{3D Gaussian Splatting.}  

3D Gaussian Splatting (3DGS) \cite{kerbl20233d} represents scenes using 3D Gaussian primitives, achieving state-of-the-art rendering quality and speed with a tile-based rasterizer. Each Gaussian is defined by its mean $\mu \in \mathbb{R}^3$ and covariance $\Sigma \in \mathbb{R}^{3 \times 3}$ with
$G(x) = e^{-\frac{1}{2}(x-\mu)^T \Sigma^{-1}(x-\mu)}$,
where $x$ is a 3D position in the scene. The rasterizer efficiently sorts and $\alpha$-blends Gaussians onto the 2D image plane \cite{zwicker2001ewa}. 
The adaptive control of Gaussians based on gradients is also critical for improving scene representation accuracy, reducing redundancy, and refining Gaussians to better match the underlying geometry and appearance.

Scaffold-GS \cite{scaffoldgs} improves 3DGS by enhancing scene structure fidelity and robustness to view-dependent effects. It uses a hierarchical 3D Gaussian representation, with anchor points encoding local scene information and generating local neural Gaussians. Each anchor, optimized with a feature vector, predicts the color, center, variance, and opacity of the neural Gaussians.

\paragraph{Neural Implicit SDFs.}

NeuS \cite{wang2021neus} integrates signed distance functions (SDFs) with NeRF’s volumetric rendering, converting SDF values into volume densities to maintain geometric accuracy. SDF values are converted to opacities using a logistic function
$
\alpha_i=\max \left(\frac{\Phi_s\left(f\left(\mathbf{x}_i\right)\right)-\Phi_s\left(f\left(\mathbf{x}_{i+1}\right)\right)}{\Phi_s\left(f\left(\mathbf{x}_i\right)\right)}, 0\right).
$
The color of a ray $r$ is also calculated by accumulating weighted colors of the sample points:
$\mathbf{C}(\mathbf{r})=\sum_{i=1}^P T_i \alpha_i \mathbf{c}_i$, and $T_i=\exp \left(-\sum_{j=1}^{i-1} \alpha_j \delta_j\right)$,
where $\delta_j$ is the interval between sampled points.
Inspired by \cite{muller2022instant}, \cite{instant-nsr-pl} introduces a multi-resolution hash grid to enhance representation power and accelerate rendering and training.

\subsection{GSDF: Dual-branch for Rendering and Reconstruction}
\label{sec:dual_branch}

To bridge the gap between rendering and geometric accuracy, we propose a novel approach that combines the strengths of Gaussian-based and SDF-based methods. By leveraging the high-quality rendering capabilities of 3DGS and the precise geometric representation of SDFs, our method aims to achieve superior results in both tasks.

As illustrated in Fig.~\ref{fig:pipeline}, our dual-branch design integrates a GS-branch and an SDF-branch. We use Scaffold-GS~\cite{scaffoldgs} and NeuS~\cite{wang2021neus} with adapted hash-encoding implementation~\cite{instant-nsr-pl} as the backbones, chosen for their effectiveness and simplicity. Importantly, our framework is versatile and can be easily adapted to incorporate future advanced methods for each branch.

\subsubsection{GS \textrightarrow{} SDF: Depth Guided Ray Sampling}
\label{sec:gs2neus}

To address the computational expenses of ray-sampling, techniques such as hierarchical sampling~\cite{mildenhall2020nerf}, occupancy grids~\cite{liu2020neural,chen2022tensorf,muller2022instant}, early stopping~\cite{muller2022instant}, and proposal networks~\cite{barron2022mip,reiser2024binary} are widely used with CUDA accelerations. When depth maps are available, either from sensors or monocular estimation, samples can be strategically placed around surface regions, which is crucial for effective optimization of the Signed Distance Field (SDF).
Unlike neural implicit SDF-based methods that rely on their own predicted SDF values for ray sampling~\cite{wang2021neus, li2023neuralangelo, rosu2023permutosdf}, we employ the GS-branch to provide surface proximity, avoiding the chicken-and-egg dilemma. Inspired by \cite{xu2023gridnerf}, where a more efficient branch provides coarse geometry guidance, we leverage depth maps from the GS-branch to refine the ray sampling range for the SDF-branch. While Gaussian primitives may be less precise, they are efficient and flexible, offering sufficient geometric clues without significant overhead.

Concretely, for a ray emitted from a camera center $\vec{o}$ in the direction $\vec{v}$, the depth value $D$ from the GS-branch is:
\begin{equation}
D = \sum_{i \in N} d_i \sigma_i \prod_{j=1}^{i-1} (1 - \sigma_j),
\label{gs_depth}
\end{equation}
where $N$ is the number of 3D Gaussians encountered, $\sigma_i$ is the opacity, and $d_i$ is the distance from the $i$-th Gaussian to the camera. For SDF-branch optimization, points are sampled around $\vec{o} + D \cdot \vec{v}$.
The sampling range adapts based on predicted SDF values $s$ at various depths:
\begin{equation}
s=\mathcal{F}_{s}(\vec{o} + D \cdot \vec{v}),
\label{sdf}
\end{equation}
where $\mathcal{F}_{s}$ is a two-layer MLP predicting the SDF value at a given position. The sampling range is defined as $r = [\vec{o} + (D - k|s|)\cdot\vec{v}, \vec{o} + (D + k|s|)\cdot\vec{v}]$. 
As shown in Fig.~\ref{fig:pipeline}, inspired from NeRF’s hierarchical sampling strategy~\cite{mildenhall2020nerf}, we use coarse and fine ranges with $k=3$ and $k=1$, respectively, we uniformly sample $M$ points along the ray within each range.

\subsubsection{SDF \textrightarrow{} GS: Geometry-aware Gaussian Density Control}
\label{sec:neus2gs}

Many previous and concurrent methods~\cite{Huang20242DGS,guedon2023sugar,Dai2024HighqualitySR} have attempted to learn scene surfaces from 3DGS by flattening 3D Gaussians along the normal direction and enforcing nearly binary opacity, treating them like surface primitives similar to mesh triangles. However, this approach often leads to degraded rendering quality and incomplete surfaces.
Our approach, instead of putting additional regularization on the 3D Gaussian primitives, enhances the distribution of Gaussian primitives using a geometry-aware density control strategy. Building on the original gradient-based density control, we leverage the zero-level set of the SDF-branch to determine the proximity of Gaussian primitives to the surface.
By querying the SDF-branch with the positions of Gaussian primitives, we are able to identify close-to-surface Gaussian primitives (i.e. with smaller absolute SDF values), allowing for more precise control over the placement and density of Gaussians.

\remove{
Prior methods have tried to learn scene surfaces from 3DGS by making 3D Gaussians flat with nearly binary opacity, 
basically treating 3D Gaussians as surface primitives, similar to mesh triangles without the water-tight restriction.
Results showed that such constraints can lead to degraded rendering quality and incomplete surfaces.
In contrast, we do not encourage the flattening of 3D Gaussian primitives, but instead improve the distribution of Gaussian primitives via a geometry-aware density control strategy. 
On top of the original gradient-based density control criteria, we additionally leverage the zero-level set of the SDF-branch to identify whether a Gaussian primitive is close to or deviates from the surface.
We query the SDF-branch with the position of Gaussian primitives.
Primitives with smaller absolute SDF values are deemed to be closer to the surface, and vice versa.
}

\noindent \textbf{Growing Operator.}
For each Gaussian primitive at position $c$, we obtain its SDF value $s = \mathcal{F}_{s}(c)$ from the SDF-branch. The growth criteria for Gaussians are then defined as:
\begin{equation}
    \epsilon_g = \nabla_g + \omega_g \mu(s),
    \label{eq:growth_strategy}
\end{equation}
where $\nabla_g$ is the averaged gradient of Gaussian primitives accumulated over $K$ training iterations, as in Scaffold-GS~\cite{scaffoldgs}. $\mu(s) = \exp\left(-s^2 / (2\sigma^2)\right)$ is a Gaussian function converting the SDF value to a positive factor, decreasing monotonically with distance from the zero level set. $\omega_g$ controls the influence of geometric guidance. New Gaussian primitives are added if $\epsilon_g$ are greater than a predefined $\tau_g$.

\noindent \textbf{Pruning Operator.}
Beyond the original opacity-based criteria, we prune Gaussian primitives far from the surface, indicated by large SDF values. The pruning criteria are:
\begin{equation}
    \epsilon_p = \sigma_a - \omega_p (1 - \mu(s)),
    \label{eq:pruning_criteria}
\end{equation}
where $\sigma_a$ is the aggregated opacity over $K$ iterations. The weight $\omega_p$ balances the contributions of transparency and SDF. Primitives with $\epsilon_p$ less than a predefined $\tau_p$ are pruned.

\subsubsection{GS \ensuremath{\leftrightarrow} SDF: Depth Guided Ray Sampling}

\label{sec:both}

To enhance both rendering and reconstruction outcomes, our framework incorporates mutual supervision between the two branches, using depth and normal maps as key geometric features to facilitate this interconnection.

For the SDF-branch, a per-view depth map $D_{s}$ is rendered using volumetric rendering principles, and a normal map $N_{s}$ is derived by rendering the gradients of the SDF volumetrically.
For the GS-branch, we compute a per-view depth map $D_{g}$ following Eq.~\ref{gs_depth}. The normal of each 3D Gaussian is determined by the direction of its smallest scaling factor, as in \cite{guedon2023sugar, chen2023neusg, cheng2023gaussian}. To render the normal map for each camera view, we accumulate the normals of the 3D Gaussians using $\alpha$-blending $N_{g} = \sum_{i \in N} n_i \sigma_i \prod_{j=1}^{i-1} (1 - \sigma_j)$,
where $n_i$ is the estimated normal of the $i$-th 3D Gaussian.

This mutual geometry supervision ensures that the depth and normal maps from both branches are consistently aligned. 
This consistent alignment is crucial for maintaining coherence between the rendering and reconstruction processes, as it helps to prevent discrepancies that could lead to artifacts and inaccuracies when performing mutual guidance. 
This consistency is especially important for achieving high-quality results in complex scenes, where precise geometric alignment can significantly enhance the visual and structural integrity of the output.

\subsection{Training Strategy and Loss Design}
\label{sec:losses}

The GS-branch is supervised by rendering losses $\mathcal{L}_1$ and $\mathcal{L}_\text{SSIM}$, which measure the difference between the rendered RGB images and ground truth images, and a volume regularization term $\mathcal{L}_{vol}$ as in~\cite{scaffoldgs}. 
The complete loss function for GS-branch is:
\begin{equation}
\mathcal{L}_{\text{g}} = \lambda_1\mathcal{L}_1 + (1-\lambda_1)\mathcal{L}_\text{SSIM} + \lambda_{vol}\mathcal{L}_{vol},
\end{equation}
where $\lambda_1$ and $\lambda_{vol}$ are weighting coefficients.
Similarly, the SDF-branch is supervised by the $\mathcal{L}_1$ rendering loss, supplemented with Eikonal and curvature penalties to ensure accurate geometry reconstruction. The loss function for the SDF-branch is:
\begin{equation}
\mathcal{L}_{\text{s}} = \mathcal{L}_1 + \lambda_\text{eik}\mathcal{L}_\text{eik} + \lambda_\text{curv}\mathcal{L}_\text{curv},
\end{equation}
where $\mathcal{L}_\text{eik}$ represents the Eikonal loss, ensuring that the gradients of the predicted SDF field are normalized, and $\mathcal{L}_\text{curv}$ denotes the curvature loss, promoting surface smoothness as described in~\cite{li2023neuralangelo,rosu2023permutosdf}. The coefficients $\lambda_\text{eik}$ and $\lambda_\text{curv}$ balance the influence of these loss terms.

The mutual geometry supervision includes depth and normal consistency losses applied to both branches. For unbounded scenes with distant backgrounds, we only apply the mutual geometric supervision to foreground regions, as these geometries are more reliable and of primary interest~\cite{instant-nsr-pl}. The mutual loss is formulated as:

\begin{equation}
\label{eq:mutual}
\mathcal{L}_{\text{mutual}} = \lambda_{d} \mathcal{L}_{d} + \lambda_{n} \mathcal{L}_{n} = \lambda_{d} \| D_{gs} - D_{s} \| + \lambda_{n} \left(1 - \frac{| N_{gs} \cdot N_{s} |}{\| N_{gs} \| \, \| N_{s} \|} \right)
\end{equation}
where $\mathcal{L}_{d}$ and $\mathcal{L}_{n}$ represent the depth and normal discrepancies between the two branches, and $\lambda_{d}$ and $\lambda_{n}$ balance their importance.
Finally, the total loss for joint learning is defined as:
\begin{equation}
   \mathcal{L} = \mathcal{L}_{\text{g}} + \mathcal{L}_{\text{s}} + \mathcal{L}_{\text{mutual}}.
\end{equation}
The hyper-parameter settings are detailed in the supplementary material.

\section{Experiment}
\label{sec:experiment}

\subsection{Experimental Setup}
\label{sec: exp_setup}
\noindent \textbf{Datasets.} 
We evaluated results using 26 real-world and synthetic scenes from various datasets: 7 from Mip-NeRF360~\cite{barron2022mip}, 2 from DeepBlending~\cite{DeepBlending2018}, 2 from Tanks\&Temples~\cite{Knapitsch2017}, and 15 from DTU~\cite{jensen2014large}, featuring a wide range of indoor, outdoor, and object-centric scenarios.

\noindent \textbf{Implementation Details.}
We implemented our dual-branch model based on 1) Scaffold-GS~\cite{scaffoldgs} and 2) an enhanced version of NeuS~\cite{wang2021neus} with a hash-grid variant~\cite{instant-nsr-pl}, following the practice of~\cite{li2023neuralangelo}. The hash grid resolution spans from $2^5$ to $2^{11}$ with 16 levels, each entry having a feature dimension of 4 and a maximum of $2^{21}$ entries per level. The coarsest 4 layers were activated initially for the DTU~\cite{aanaes2016large}, and 8 layers for other datasets, with finer levels added every 2$k$ iterations.
We trained the GS-branch for 15k iterations, followed by joint training of both branches for 30k iterations. The SDF-branch was warmed up for 2k iterations on the DTU and 5k on other datasets without depth-guided ray sampling.
Note that our system is adaptable with various existing or future rendering and reconstruction models.

\noindent \textbf{Evaluations.}
We evaluated our method against state-of-the-art rendering and reconstruction approaches. 
For rendering comparison, we compared with Scaffold-GS~\cite{scaffoldgs}, 3D-GS~\cite{kerbl20233d}, NeuS~\cite{wang2021neus}, 2D-GS~\cite{Huang20242DGS}, and SuGaR~\cite{guedon2023sugar} using PSNR, SSIM~\cite{1284395}, and LPIPS~\cite{Zhang_2018_CVPR} for quantitative comparisons. We trained 3D-GS and Scaffold-GS for 45$k$ iterations to align with our configurations.
For reconstruction, we compared with NeuS~\cite{wang2021neus}, Instant-NSR~\cite{instant-nsr-pl} (our representative SDF-branch), SuGaR~\cite{guedon2023sugar} and 2D-GS~\cite{Huang20242DGS}. We use Chamfer distance for quantitative evaluation. 
For all datasets, we used $1/8$ of the images as test sets and the other $7/8$ as training sets.
All experiments were conducted on a single NVIDIA A100 GPU with 80G memory.

\subsection{Results Analysis}

\begin{table}[t!]
\centering
\caption{\textbf{Rendering and reconstruction comparisons against baselines} over four benchmark scenes. 3D-GS~\cite{kerbl20233d}, Scaffold-GS~\cite{scaffoldgs}, 2D-GS~\cite{Huang20242DGS}, SuGaR~\cite{guedon2023sugar} and GSDF initialized the Gaussian primitives with COLMAP~\cite{schonberger2016structure} sparse points.
}
\label{tab:quality_update}
\resizebox{\linewidth}{!}{
\begin{tabular}{cc|cccc|ccc|ccc|ccc}
\toprule
\multicolumn{2}{c|}{Dataset} & \multicolumn{4}{c|}{DTU~\cite{aanaes2016large}} & \multicolumn{3}{c|}{Tanks\&Temples~\cite{Knapitsch2017}} & \multicolumn{3}{c|}{Mip-NeRF360~\cite{barron2022mip}} & \multicolumn{3}{c}{Deep Blending~\cite{DeepBlending2018}} \\ \midrule
& & \multicolumn{1}{c|}{reconstruction} & \multicolumn{3}{c|}{rendering} & \multicolumn{3}{c|}{rendering} & \multicolumn{3}{c|}{rendering} & \multicolumn{3}{c}{rendering} \\ 
\multicolumn{2}{c|}{\multirow{-2}{*}{Method \& Metrics}} & \multicolumn{1}{c|}{CD \(\downarrow\)} & PSNR \(\uparrow\) & SSIM \(\uparrow\) & LPIPS \(\downarrow\) & PSNR \(\uparrow\) & SSIM \(\uparrow\) & LPIPS \(\downarrow\) & PSNR \(\uparrow\) & SSIM \(\uparrow\) & LPIPS \(\downarrow\) & PSNR \(\uparrow\) & SSIM \(\uparrow\) & LPIPS \(\downarrow\) \\ \midrule
  & NeuS~\cite{wang2021neus} & \cellcolor{tabsecond}0.823 & 31.93 & 0.912 & 0.165 & - & - & - & - & - & - & - & - & - \\
\multirow{-2}{*}{SDF-Based}  & Instant-NSR~\cite{instant-nsr-pl} & 1.809 & 26.34 & 0.846 & 0.230 & 19.36 & 0.641 & 0.495 & 21.87 & 0.612 & 0.509 & 24.00 & 0.795 & 0.462 \\ \midrule
  & 3D-GS~\cite{kerbl20233d} & 4.034 & 32.91 & 0.943 & \cellcolor{tabfirst}0.092 & \cellcolor{tabthird}26.73 & \cellcolor{tabthird}0.858 & \cellcolor{tabthird}0.210 & \cellcolor{tabthird}28.89 & \cellcolor{tabthird}0.857 & \cellcolor{tabthird}0.209 & 29.44 & \cellcolor{tabthird}0.899 & \cellcolor{tabthird}0.248 \\
  & Scaffold-GS~\cite{scaffoldgs}& 5.988& \cellcolor{tabthird}33.05 & \cellcolor{tabthird}0.946 & 0.100 & \cellcolor{tabsecond}27.29 & \cellcolor{tabsecond}0.869 & \cellcolor{tabsecond}0.182 & \cellcolor{tabsecond}29.34 & \cellcolor{tabsecond}0.863 & \cellcolor{tabsecond}0.200 & \cellcolor{tabsecond}30.37 & \cellcolor{tabsecond}0.908 & \cellcolor{tabsecond}0.238 \\
  & 2D-GS~\cite{Huang20242DGS}& \cellcolor{tabthird}0.928 & \cellcolor{tabsecond}33.39 & \cellcolor{tabsecond}0.947 & 0.105 & 25.75 & 0.840 & 0.246 & 28.08 & 0.843 & 0.240 & 29.27 & 0.897 & 0.261 \\
\multirow{-4}{*}{GS-Based} & SuGaR~\cite{guedon2023sugar}& 1.239 & 32.76 & 0.942 & \cellcolor{tabthird}0.094 & 24.68 & 0.827 & 0.230 & 27.40 & 0.817 & 0.260 & \cellcolor{tabthird}29.53 & 0.895 & 0.265 \\ \midrule
Hybrid & \textbf{GSDF} & \cellcolor{tabfirst}0.802 & \cellcolor{tabfirst}33.65 & \cellcolor{tabfirst}0.948 & \cellcolor{tabfirst}0.092 & \cellcolor{tabfirst}27.37 & \cellcolor{tabfirst}0.875 & \cellcolor{tabfirst}0.156 & \cellcolor{tabfirst}29.38 & \cellcolor{tabfirst}0.865 & \cellcolor{tabfirst}0.185 & \cellcolor{tabfirst}30.38 & \cellcolor{tabfirst}0.909 & \cellcolor{tabfirst}0.223 \\

\bottomrule
\end{tabular}}
\end{table}

\begin{table}[t!]
\centering
\caption{\textbf{Rendering comparisons with random initialization against baselines} over three benchmark scenes. Scaffold-GS (rand)~\cite{scaffoldgs}, 2D-GS (rand)~\cite{Huang20242DGS} and GSDF (rand) initialized the Gaussian primitives with random points. 
}
\label{tab:quality_random}
\resizebox{\linewidth}{!}{
\begin{tabular}{cc|ccc|ccc|ccc}
\toprule
\multicolumn{2}{c|}{Dataset} & \multicolumn{3}{c|}{Mip-NeRF360~\cite{barron2022mip}} & \multicolumn{3}{c|}{Tanks\&Temples~\cite{Knapitsch2017}} & \multicolumn{3}{c}{Deep Blending~\cite{DeepBlending2018}} \\
\multicolumn{2}{c|}{Method \& Metrics}  & PSNR \(\uparrow\) & SSIM \(\uparrow\) & LPIPS \(\downarrow\) & PSNR \(\uparrow\) & SSIM \(\uparrow\) & LPIPS \(\downarrow\) & PSNR \(\uparrow\) & SSIM \(\uparrow\) & LPIPS \(\downarrow\) \\
\midrule

\multicolumn{2}{c|}{2D-GS (rand)} & 26.94 & 0.793 & 0.300 & 25.16 & 0.817 & 0.275  &  28.76  & 0.892 & 0.274 \\
\multicolumn{2}{c|}{Scaffold-GS (rand)} & \underline{27.84} & \underline{0.817} & \underline{0.259} & \underline{26.37} & \underline{0.838} & \underline{0.230}  & \underline{29.12}  & \underline{0.895} & \underline{0.260} \\

\multicolumn{2}{c|}{\textbf{GSDF} (rand)} & \textbf{28.05} & \textbf{0.830} & \textbf{0.229} & \textbf{26.55} & \textbf{0.852} & \textbf{0.197} & \textbf{29.57} & \textbf{0.899} & \textbf{0.244} \\
\bottomrule
\end{tabular}}
\end{table}

\begin{figure}[t!]
\centering
\vspace{-10pt}
\includegraphics[width=\linewidth]
{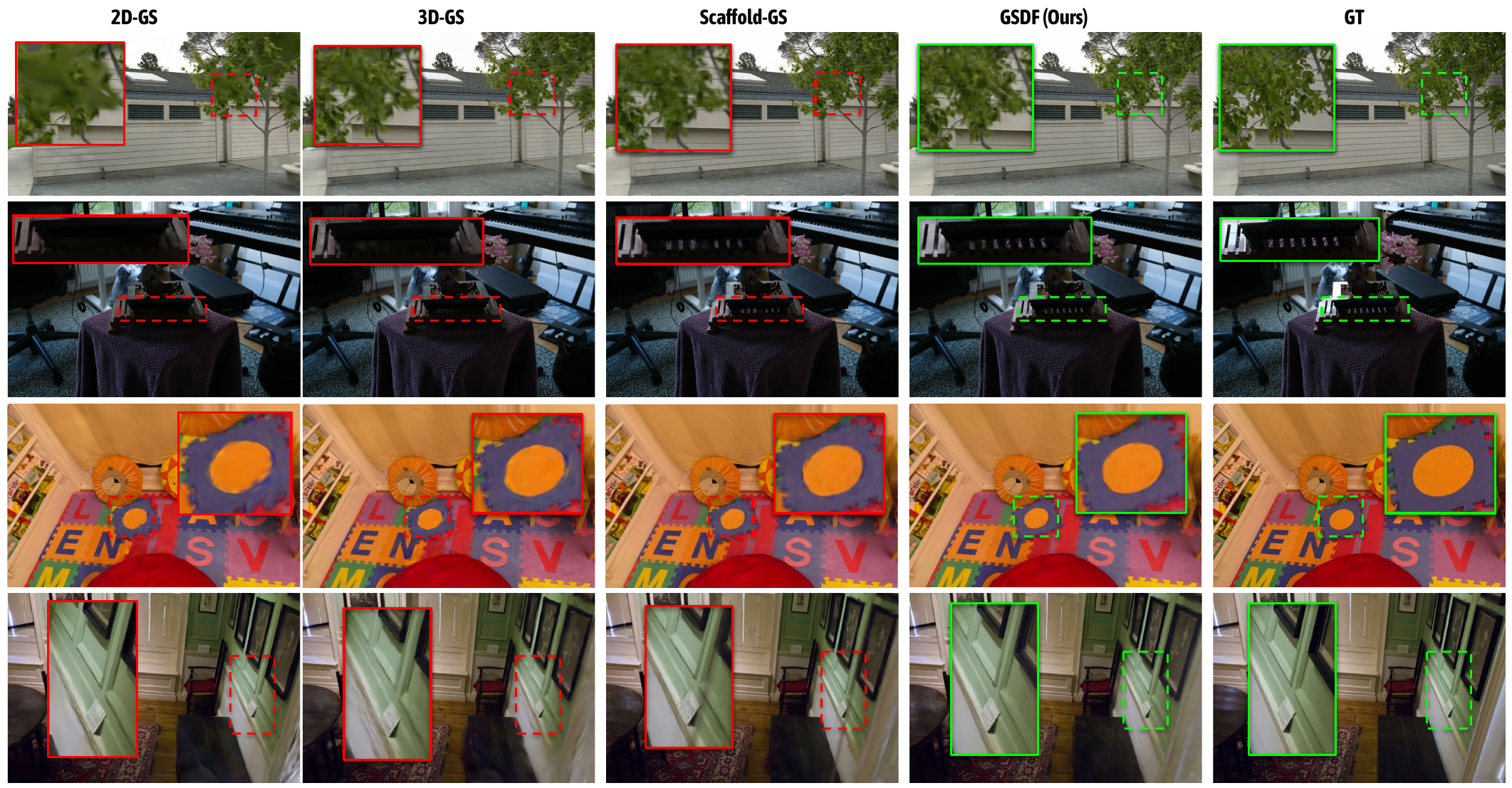}
\caption{\textbf{Qualitative comparisons} of GSDF against popular Gaussian-based baselines~\cite{kerbl20233d, scaffoldgs, Huang20242DGS} across diverse 3D scene datasets~\cite{Knapitsch2017, barron2022mip, DeepBlending2018}. 
As highlighted, GSDF excels in modeling delicate geometries (1st \& 2nd rows) and handling texture-less and sparsely observed regions (3rd \& 4th rows), which are commonly presented in larger scenes where baseline approaches struggle to address.
}
\label{fig:rendering_comparison}
\end{figure}

\begin{figure}[t!]

  \centering
   \includegraphics[width=1\linewidth]{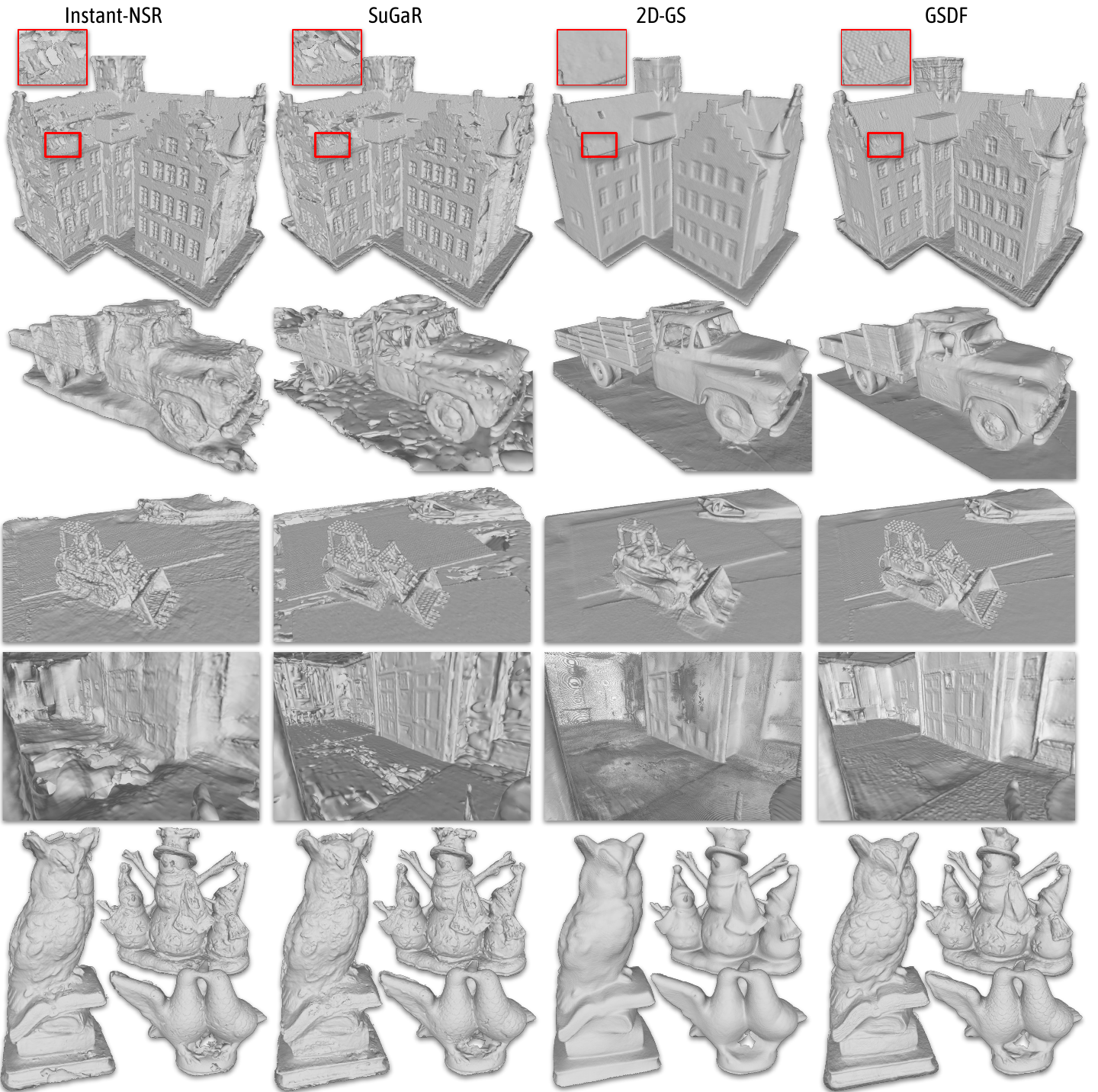}
   
   \caption{\textbf{Reconstruction Comparison}. We visualize the reconstructed meshes from Instant-NSR~\cite{instant-nsr-pl}~(our SDF-branch), SuGaR~\cite{guedon2023sugar}, 2D-GS~\cite{Huang20242DGS}, and Ours. 
   }

   \label{fig:mesh_comparison}
\end{figure}

\subsubsection{Rendering Comparisons} 
Our GSDF retained high rendering quality compared to state-of-the-art 3D-GS and SDF-based methods, as shown in Table~\ref{tab:quality_update}. 
Quantitative metrics against 3D-GS~\cite{kerbl20233d}, 2D-GS~\cite{Huang20242DGS}, and Scaffold-GS~\cite{scaffoldgs} show that GSDF consistently outperformed these methods across all four datasets, showcasing improvements in all metrics. 
Notably, the prominent improvements in the LPIPS metric indicate that our method effectively captures high-frequency scene details and renders perceptually superior results compared to baselines. 

\noindent \textbf{Enchanced Detail and Fidelity.}
Notably, our method excelled in achieving \emph{high rendering quality} in texture-less areas, as showcased in Fig.\ref{fig:rendering_comparison}, which is aligned with our design motivation for geometry-aware density control.
Specifically, in texture-less areas where vanilla 3D Gaussians struggled due to small accumulated gradients, our method overcame this limitation by growing anchors in surface regions which brings enhanced accuracy and scene details. 
\noindent \textbf{Robustness with Noisy Gaussians.}
To test the robustness, we experimented with randomly initialized Gaussian primitives on 2D-GS, Scaffold-GS, and our method. Quantitative results in Table~\ref{tab:quality_random} highlight the advantages of our geometric guidance, demonstrating superior performance and stability even with random input. 
Further visual results are provided in the supplementary material. 
\subsubsection{Reconstruction Comparisons} 
We quantitatively evaluated the reconstruction ability on the DTU dataset, where our method achieved the best Chamfer distance as shown in Table~\ref{tab:quality_update}. %

\noindent \textbf{Enhanced Geometry Accuracy and Completeness.}
As illustrated in Fig.\ref{fig:mesh_comparison}, our method reconstructed more complete and detailed meshes compared to baseline methods. Our approach effectively bypassed local minima, preventing holes in the meshes. Notably, our method outperformed Instant-NSR, delivering smooth, continuous meshes with well-preserved high-frequency details. In contrast, meshes extracted from SuGaR~\cite{guedon2023sugar} were non-manifold with broken topological relationships. 2D-GS~\cite{Huang20242DGS} extracts mesh from fused depth, which leads to over-smooth geometry.

\noindent \textbf{Accerelated Convergence and Rendering.}
Benefited from the GS-branch, our method optimized the SDF field significantly faster  in terms of training iterations than previous methods with accurate sample placements, which often required slow and exhaustive training and rendering. 
For instance, NeuS~\cite{wang2021neus} required about 8 hours of optimization on a single GPU for the DTU dataset, whereas our method achieved comparable or better results within just 2 hours on a single GPU.

\subsection{Ablation Studies}
\label{Sec:ablation}
\begin{table}[t!]
\centering
\caption{\textbf{Quantitative Results on Ablation Studies.} We separately listed the rendering metrics for each ablation described in Sec.~\ref{Sec:ablation}.
}
\label{tab:ablation}
\resizebox{1\linewidth}{!}{
\begin{tabular}{c|ccc|ccc|ccc}
\toprule
Dataset & \multicolumn{3}{c|}{Mip-NeRF360\cite{barron2022mip}} & \multicolumn{3}{c|}{Tanks\&Temples\cite{Knapitsch2017}} & \multicolumn{3}{c}{Deep Blending~\cite{DeepBlending2018}} \\
\begin{tabular}{c|c} Method & Metrics \end{tabular}  & PSNR \(\uparrow\) & SSIM \(\uparrow\) & LPIPS \(\downarrow\) & PSNR \(\uparrow\) & SSIM \(\uparrow\) & LPIPS \(\downarrow\) & PSNR \(\uparrow\) & SSIM \(\uparrow\) & LPIPS \(\downarrow\) \\
\midrule
\text{GSDF (Full)} & \textbf{29.38} & \textbf{0.865} & \textbf{0.185} & \textbf{27.37} & \textbf{0.875} & \textbf{0.156} & \textbf{30.38} & \textbf{0.909} & \textbf{0.223} \\ \hline

w/o  geometric supervision & 29.27 & 0.862 & 0.196 & 27.18 & 0.865 & 0.191 & 30.21 & 0.907 & 0.234 \\

 w/o depth-guided sampling & 29.26 & 0.863 & 0.196 & 27.30 &0.873 &0.159 & 30.17 & 0.908 & 0.228 \\ 
 
$\text{w/o geometry-aware densification}$ & 29.29 & 0.863 & 0.196 & 27.29 &0.870 &0.179 & 30.25 & 0.908 & 0.236 \\

\bottomrule
\end{tabular}}
\end{table}

\begin{figure}[t]

  \centering
   \includegraphics[width=1\linewidth]{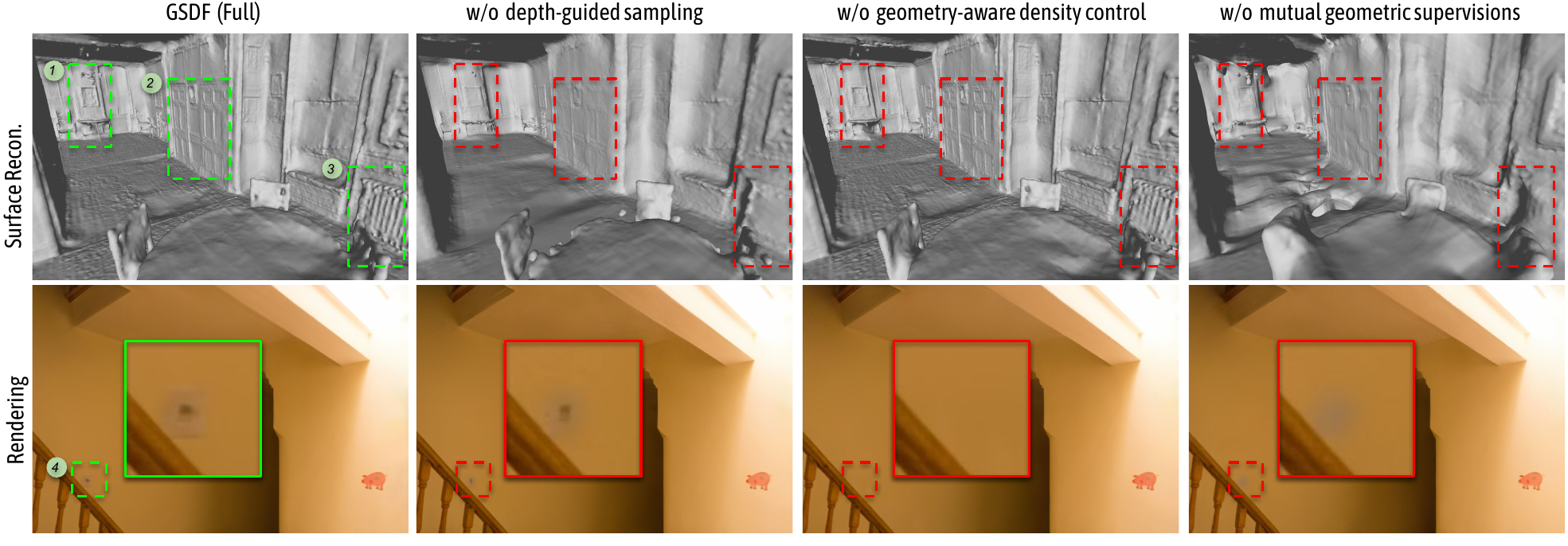}
   
   \caption{\textbf{Ablation results}. Visualizations of reconstructed meshes and rendered images from 1) our full method,  2) ours w/o depth-guided ray sampling, 3) ours w/o geometry-aware density control, and 4) ours w/o geometric supervision. We highlight the degradation of quality using numbered patches.
   }

   \label{fig:ablation_mesh}
\end{figure}

In this section, we examine the effectiveness of each individual module. Quantitative metrics and qualitative visualizations are provided in Table~\ref{tab:ablation} and Fig.~\ref{fig:ablation_mesh}.

\noindent \textbf{Depth-Guided Ray Sampling.}
To evaluate the effectiveness of our depth-guided ray sampling detailed in Sec~\ref{sec:dual_branch}, we conducted an ablation on depth-guided ray sampling using the original stratified ray sampling approach~\cite{mildenhall2020nerf}. The results show that this ablated setting produced overly smoothed surfaces, failing to capture finer geometry details such as the heater and doors (Fig.~\ref{fig:ablation_mesh}, column 2, patches 2 \& 3). 
The resulting less accurate SDF also impacted rendering quality, making details like the sticker on the wall appear more blurred compared to our full model.

\noindent \textbf{Geometry-aware Gaussian Density Control.} 
We replaced our geometry-aware Gaussian density control with the pruning and growing strategy from Scaffold-GS~\cite{scaffoldgs} to evaluate its efficacy. As shown in Fig.~\ref{fig:ablation_mesh} (3rd column), this change led to missed details like the sticker on the wall (patch 4) due to small accumulated gradients on the texture-less surface. The reconstruction results also showed the absence of thin objects such as table legs and the chandelier (patch 1), highlighting the limitations of the vanilla strategy.

\noindent \textbf{Mutual Geometric Supervision.} 
Lastly, we ablated the proposed mutual geometric supervision by setting both $\lambda_d$ and $\lambda_n$ to 0 (Eq.~\ref{eq:mutual}). This resulted in a significant decay in surface quality and omission of details in rendering. These findings indicate that neural rendering is more tolerant of deviations than neural surface reconstruction. Without explicitly aligning Gaussians with SDF-derived geometry during optimization, the two branches can diverge, leading to sub-optimal results for both.
%
\subsection{Limitations and Future Works}
\label{sec:limitations}

Currently, our method is not tailored to handle challenging scenes with reflections and intense lighting variations, such as those found in indoor environments. 
However, we have observed that employing more structured and surface-aligned Gaussian primitives holds promise in capturing such view-dependent appearance changes with improved scene geometry.
Our framework requires more memory usage because it simultaneously considers two representations.
Furthermore, the performance of our SDF-branch significantly lags behind the GS-branch, leading to extended training durations compared to Scaffold-GS~\cite{scaffoldgs} only. 
Hence, improving the efficiency of the MLP-based SDF-branch is a crucial direction for future research.
Still, training time is not the highest priority compared to inference, where primitives are stored and can be accessed efficiently.

\section{Conclusion}
\label{sec:conclusion}

In this work, we introduced a dual-branch framework that leverages the strengths of both 3D-GS and SDF, showcasing its potential to achieve both enhanced rendering and reconstruction quality.
The inherent differences in two representations, rendering approaches, and supervision loss pose a challenge to the seamless integration. 
We therefore integrate a bidirectional mutual guidance approach during the training to circumvent these restrictions.
Three types of guidance have been introduced and validated in our framework, namely: 1) depth guided sampling (GS$\rightarrow$SDF), 2) geometry-aware Gaussian density control (SDF$\rightarrow$GS); and 3) mutual geometry supervision (GS$\leftrightarrow$SDF). Our extensive results demonstrate the efficiency and joint performance improvement on both tasks. 
As the two branches maintain their original architectures, we keep their efficiency during inference, allowing room for potential enhancements by substituting each branch with more advanced models in the future. 
%
We envision our model benefiting applications demanding high-quality rendering and geometry, including embodied environments, physical simulation, and immersive VR experiences.

\section{Acknowledgements}
This work is funded in part by the National Key R\&D Program of China (2022ZD0160201), and Shanghai Artificial Intelligence Laboratory.


\begin{thebibliography}{10}

  \bibitem{aanaes2016large}
  Henrik Aan{\ae}s, Rasmus~Ramsb{\o}l Jensen, George Vogiatzis, Engin Tola, and Anders~Bjorholm Dahl.
  \newblock Large-scale data for multiple-view stereopsis.
  \newblock {\em International Journal of Computer Vision}, pages 1--16, 2016.
  
  \bibitem{aliev2020neural}
  Kara-Ali Aliev, Artem Sevastopolsky, Maria Kolos, Dmitry Ulyanov, and Victor Lempitsky.
  \newblock Neural point-based graphics.
  \newblock In {\em Computer Vision--ECCV 2020: 16th European Conference, Glasgow, UK, August 23--28, 2020, Proceedings, Part XXII 16}, pages 696--712. Springer, 2020.
  
  \bibitem{barron2022mip}
  Jonathan~T Barron, Ben Mildenhall, Dor Verbin, Pratul~P Srinivasan, and Peter Hedman.
  \newblock Mip-nerf 360: Unbounded anti-aliased neural radiance fields.
  \newblock In {\em Proceedings of the IEEE/CVF Conference on Computer Vision and Pattern Recognition}, pages 5470--5479, 2022.
  
  \bibitem{chen2022tensorf}
  Anpei Chen, Zexiang Xu, Andreas Geiger, Jingyi Yu, and Hao Su.
  \newblock {TensoRF}: Tensorial radiance fields.
  \newblock In {\em ECCV}, 2022.
  
  \bibitem{chen2023neusg}
  Hanlin Chen, Chen Li, and Gim~Hee Lee.
  \newblock Neusg: Neural implicit surface reconstruction with 3d gaussian splatting guidance.
  \newblock {\em arXiv preprint arXiv:2312.00846}, 2023.
  
  \bibitem{cheng2023gaussian}
  Kai Cheng, Xiaoxiao Long, Kaizhi Yang, Yao Yao, Wei Yin, Yuexin Ma, Wenping Wang, and Xuejin Chen.
  \newblock Gaussianpro: 3d gaussian splatting with progressive propagation.
  \newblock {\em arXiv preprint arXiv:}, 2024.
  
  \bibitem{Dai2024HighqualitySR}
  Pinxuan Dai, Jiamin Xu, Wenxiang Xie, Xinguo Liu, Huamin Wang, and Weiwei Xu.
  \newblock High-quality surface reconstruction using gaussian surfels.
  \newblock In Andres Burbano, Denis Zorin, and Wojciech Jarosz, editors, {\em {ACM} {SIGGRAPH} 2024 Conference Papers, {SIGGRAPH} 2024, Denver, CO, USA, 27 July 2024- 1 August 2024}, page~22. {ACM}, 2024.
  
  \bibitem{feng2024gaussian}
  Yutao Feng, Xiang Feng, Yintong Shang, Ying Jiang, Chang Yu, Zeshun Zong, Tianjia Shao, Hongzhi Wu, Kun Zhou, Chenfanfu Jiang, et~al.
  \newblock Gaussian splashing: Dynamic fluid synthesis with gaussian splatting.
  \newblock {\em arXiv preprint arXiv:2401.15318}, 2024.
  
  \bibitem{fridovich2022plenoxels}
  Sara Fridovich-Keil, Alex Yu, Matthew Tancik, Qinhong Chen, Benjamin Recht, and Angjoo Kanazawa.
  \newblock Plenoxels: Radiance fields without neural networks.
  \newblock In {\em Proceedings of the IEEE/CVF Conference on Computer Vision and Pattern Recognition}, pages 5501--5510, 2022.
  
  \bibitem{guedon2023sugar}
  Antoine Gu{\'e}don and Vincent Lepetit.
  \newblock Sugar: Surface-aligned gaussian splatting for efficient 3d mesh reconstruction and high-quality mesh rendering.
  \newblock {\em arXiv preprint arXiv:2311.12775}, 2023.
  
  \bibitem{instant-nsr-pl}
  Yuan-Chen Guo.
  \newblock Instant neural surface reconstruction, 2022.
  \newblock https://github.com/bennyguo/instant-nsr-pl.
  
  \bibitem{DeepBlending2018}
  Peter Hedman, Julien Philip, True Price, Jan{-}Michael Frahm, George Drettakis, and Gabriel~J. Brostow.
  \newblock Deep blending for free-viewpoint image-based rendering.
  \newblock {\em {ACM} Trans. Graph.}, 37(6):257, 2018.
  
  \bibitem{Huang20242DGS}
  Binbin Huang, Zehao Yu, Anpei Chen, Andreas Geiger, and Shenghua Gao.
  \newblock 2d gaussian splatting for geometrically accurate radiance fields.
  \newblock {\em ArXiv}, abs/2403.17888, 2024.
  
  \bibitem{jensen2014large}
  Rasmus Jensen, Anders Dahl, George Vogiatzis, Engin Tola, and Henrik Aan{\ae}s.
  \newblock Large scale multi-view stereopsis evaluation.
  \newblock In {\em Proceedings of the IEEE conference on computer vision and pattern recognition}, pages 406--413, 2014.
  
  \bibitem{jiang2024vr}
  Ying Jiang, Chang Yu, Tianyi Xie, Xuan Li, Yutao Feng, Huamin Wang, Minchen Li, Henry Lau, Feng Gao, Yin Yang, et~al.
  \newblock Vr-gs: A physical dynamics-aware interactive gaussian splatting system in virtual reality.
  \newblock {\em arXiv preprint arXiv:2401.16663}, 2024.
  
  \bibitem{keetha2023splatam}
  Nikhil Keetha, Jay Karhade, Krishna~Murthy Jatavallabhula, Gengshan Yang, Sebastian Scherer, Deva Ramanan, and Jonathon Luiten.
  \newblock Splatam: Splat, track \& map 3d gaussians for dense rgb-d slam.
  \newblock {\em arXiv preprint arXiv:2312.02126}, 2023.
  
  \bibitem{kerbl20233d}
  Bernhard Kerbl, Georgios Kopanas, Thomas Leimk{\"u}hler, and George Drettakis.
  \newblock 3d gaussian splatting for real-time radiance field rendering.
  \newblock {\em ACM Transactions on Graphics}, 42(4), 2023.
  
  \bibitem{Knapitsch2017}
  Arno Knapitsch, Jaesik Park, Qian-Yi Zhou, and Vladlen Koltun.
  \newblock Tanks and temples: Benchmarking large-scale scene reconstruction.
  \newblock {\em ACM Transactions on Graphics}, 36(4), 2017.
  
  \bibitem{li2023neuralangelo}
  Zhaoshuo Li, Thomas M{\"u}ller, Alex Evans, Russell~H Taylor, Mathias Unberath, Ming-Yu Liu, and Chen-Hsuan Lin.
  \newblock Neuralangelo: High-fidelity neural surface reconstruction.
  \newblock In {\em Proceedings of the IEEE/CVF Conference on Computer Vision and Pattern Recognition}, pages 8456--8465, 2023.
  
  \bibitem{liu2020neural}
  Lingjie Liu, Jiatao Gu, Kyaw Zaw~Lin, Tat-Seng Chua, and Christian Theobalt.
  \newblock Neural sparse voxel fields.
  \newblock {\em Advances in Neural Information Processing Systems}, 33:15651--15663, 2020.
  
  \bibitem{scaffoldgs}
  Tao Lu, Mulin Yu, Linning Xu, Yuanbo Xiangli, Limin Wang, Dahua Lin, and Bo~Dai.
  \newblock Scaffold-gs: Structured 3d gaussians for view-adaptive rendering, 2023.
  
  \bibitem{lyu20243dgsr}
  Xiaoyang Lyu, Yang-Tian Sun, Yi-Hua Huang, Xiuzhe Wu, Ziyi Yang, Yilun Chen, Jiangmiao Pang, and Xiaojuan Qi.
  \newblock 3dgsr: Implicit surface reconstruction with 3d gaussian splatting.
  \newblock {\em arXiv preprint arXiv:2404.00409}, 2024.
  
  \bibitem{mildenhall2020nerf}
  Ben Mildenhall, Pratul~P. Srinivasan, Matthew Tancik, Jonathan~T. Barron, Ravi Ramamoorthi, and Ren Ng.
  \newblock Nerf: Representing scenes as neural radiance fields for view synthesis.
  \newblock In {\em ECCV}, 2020.
  
  \bibitem{muller2022instant}
  Thomas M{\"u}ller, Alex Evans, Christoph Schied, and Alexander Keller.
  \newblock Instant neural graphics primitives with a multiresolution hash encoding.
  \newblock {\em ACM Transactions on Graphics (ToG)}, 41(4):1--15, 2022.
  
  \bibitem{oechsle2021unisurf}
  Michael Oechsle, Songyou Peng, and Andreas Geiger.
  \newblock Unisurf: Unifying neural implicit surfaces and radiance fields for multi-view reconstruction.
  \newblock In {\em Proceedings of the IEEE/CVF International Conference on Computer Vision}, pages 5589--5599, 2021.
  
  \bibitem{reiser2024binary}
  Christian Reiser, Stephan Garbin, Pratul~P Srinivasan, Dor Verbin, Richard Szeliski, Ben Mildenhall, Jonathan~T Barron, Peter Hedman, and Andreas Geiger.
  \newblock Binary opacity grids: Capturing fine geometric detail for mesh-based view synthesis.
  \newblock {\em arXiv preprint arXiv:2402.12377}, 2024.
  
  \bibitem{rosinol2023nerf}
  Antoni Rosinol, John~J Leonard, and Luca Carlone.
  \newblock Nerf-slam: Real-time dense monocular slam with neural radiance fields.
  \newblock In {\em 2023 IEEE/RSJ International Conference on Intelligent Robots and Systems (IROS)}, pages 3437--3444. IEEE, 2023.
  
  \bibitem{rosu2023permutosdf}
  Radu~Alexandru Rosu and Sven Behnke.
  \newblock Permutosdf: Fast multi-view reconstruction with implicit surfaces using permutohedral lattices.
  \newblock In {\em Proceedings of the IEEE/CVF Conference on Computer Vision and Pattern Recognition}, pages 8466--8475, 2023.
  
  \bibitem{schonberger2016structure}
  Johannes~L Schonberger and Jan-Michael Frahm.
  \newblock Structure-from-motion revisited.
  \newblock In {\em Proceedings of the IEEE conference on computer vision and pattern recognition}, pages 4104--4113, 2016.
  
  \bibitem{turki2023hybridnerf}
  Haithem Turki, Vasu Agrawal, Samuel~Rota Bul{\`o}, Lorenzo Porzi, Peter Kontschieder, Deva Ramanan, Michael Zollh{\"o}fer, and Christian Richardt.
  \newblock Hybridnerf: Efficient neural rendering via adaptive volumetric surfaces.
  \newblock {\em arXiv preprint arXiv:2312.03160}, 2023.
  
  \bibitem{wang2021neus}
  Peng Wang, Lingjie Liu, Yuan Liu, Christian Theobalt, Taku Komura, and Wenping Wang.
  \newblock Neus: Learning neural implicit surfaces by volume rendering for multi-view reconstruction.
  \newblock {\em arXiv preprint arXiv:2106.10689}, 2021.
  
  \bibitem{Wang2022NeuS2FL}
  Yiming Wang, Qin Han, Marc Habermann, Kostas Daniilidis, Christian Theobalt, and Lingjie Liu.
  \newblock Neus2: Fast learning of neural implicit surfaces for multi-view reconstruction.
  \newblock {\em 2023 IEEE/CVF International Conference on Computer Vision (ICCV)}, pages 3272--3283, 2022.
  
  \bibitem{1284395}
  Zhou Wang, A.C. Bovik, H.R. Sheikh, and E.P. Simoncelli.
  \newblock Image quality assessment: from error visibility to structural similarity.
  \newblock {\em IEEE Transactions on Image Processing}, 13(4):600--612, 2004.
  
  \bibitem{wang2023adaptive}
  Zian Wang, Tianchang Shen, Merlin Nimier-David, Nicholas Sharp, Jun Gao, Alexander Keller, Sanja Fidler, Thomas M{\"u}ller, and Zan Gojcic.
  \newblock Adaptive shells for efficient neural radiance field rendering.
  \newblock {\em arXiv preprint arXiv:2311.10091}, 2023.
  
  \bibitem{xie2023physgaussian}
  Tianyi Xie, Zeshun Zong, Yuxin Qiu, Xuan Li, Yutao Feng, Yin Yang, and Chenfanfu Jiang.
  \newblock Physgaussian: Physics-integrated 3d gaussians for generative dynamics.
  \newblock {\em arXiv preprint arXiv:2311.12198}, 2023.
  
  \bibitem{xu2023vr}
  Linning Xu, Vasu Agrawal, William Laney, Tony Garcia, Aayush Bansal, Changil Kim, Samuel Rota~Bul{\`o}, Lorenzo Porzi, Peter Kontschieder, Alja{\v{z}} Bo{\v{z}}i{\v{c}}, et~al.
  \newblock Vr-nerf: High-fidelity virtualized walkable spaces.
  \newblock In {\em SIGGRAPH Asia 2023 Conference Papers}, pages 1--12, 2023.
  
  \bibitem{xu2023gridnerf}
  Linning Xu, Yuanbo Xiangli, Sida Peng, Xingang Pan, Nanxuan Zhao, Christian Theobalt, Bo~Dai, and Dahua Lin.
  \newblock Grid-guided neural radiance fields for large urban scenes.
  \newblock In {\em CVPR}, 2023.
  
  \bibitem{xu2022point}
  Qiangeng Xu, Zexiang Xu, Julien Philip, Sai Bi, Zhixin Shu, Kalyan Sunkavalli, and Ulrich Neumann.
  \newblock Point-nerf: Point-based neural radiance fields.
  \newblock In {\em Proceedings of the IEEE/CVF Conference on Computer Vision and Pattern Recognition}, pages 5438--5448, 2022.
  
  \bibitem{yariv2021volume}
  Lior Yariv, Jiatao Gu, Yoni Kasten, and Yaron Lipman.
  \newblock Volume rendering of neural implicit surfaces.
  \newblock {\em Advances in Neural Information Processing Systems}, 34:4805--4815, 2021.
  
  \bibitem{yariv2023bakedsdf}
  Lior Yariv, Peter Hedman, Christian Reiser, Dor Verbin, Pratul~P Srinivasan, Richard Szeliski, Jonathan~T Barron, and Ben Mildenhall.
  \newblock Bakedsdf: Meshing neural sdfs for real-time view synthesis.
  \newblock {\em arXiv preprint arXiv:2302.14859}, 2023.
  
  \bibitem{yifan2019differentiable}
  Wang Yifan, Felice Serena, Shihao Wu, Cengiz {\"O}ztireli, and Olga Sorkine-Hornung.
  \newblock Differentiable surface splatting for point-based geometry processing.
  \newblock {\em ACM Transactions on Graphics (TOG)}, 38(6):1--14, 2019.
  
  \bibitem{Zhang_2018_CVPR}
  Richard Zhang, Phillip Isola, Alexei~A. Efros, Eli Shechtman, and Oliver Wang.
  \newblock The unreasonable effectiveness of deep features as a perceptual metric.
  \newblock In {\em Proceedings of the IEEE Conference on Computer Vision and Pattern Recognition (CVPR)}, June 2018.
  
  \bibitem{zwicker2001ewa}
  Matthias Zwicker, Hanspeter Pfister, Jeroen Van~Baar, and Markus Gross.
  \newblock Ewa volume splatting.
  \newblock In {\em Proceedings Visualization, 2001. VIS'01.}, pages 29--538. IEEE, 2001.
  
  \end{thebibliography}


\appendix

\newpage
\section{Supplementary Material}

The following sections are organized as follows: 
1) The first section elaborates on the implementation details of \textit{GSDF}, covering hyper-parameters and curvature loss.
2) We then show additional experimental results. 
3) Lastly, we also delve into limitations and future directions.
More results and video demos can be found in supplementary webpage.

\subsection{Implementation details}
\label{sec:impl_details}
\subsubsection{Configurations.} 
Below, we enumerate the hyper-parameters used in our experiments:
\begin{itemize}
  \item 
  The variance $\sigma^2$ for the Gaussian function in Eq.~\ref{eq:growth_strategy} and~\ref{eq:pruning_criteria} is set to 0.005.
  \item For the rendering loss discussed in Sec.~\ref{sec:losses}, we set $\lambda_1 = 0.2$ and $\lambda_{vol} = 0.01$, in consistency with the configurations specified in Scaffold-GS~\cite{scaffoldgs}.
  \item 
  For the SDF-branch, we set $\lambda_{eik} = 0.1$ and implement an adaptive scheme for $\lambda_{curv}$. Specifically, $\lambda_{curv}$ increases linearly from $0$ to $1$ ($0.5$ on DTU) over the first $2000$ iterations, after which it remains at $0.05$ ($0.01$ on DTU) for subsequent iterations. This strategy is based on our observation that increasing the weight of the curvature loss significantly encourages the convergence of the SDF to a geometric outline.
  
  \item  For the mutual geometry loss, we typically assign $\lambda_d = 0.5$ and $\lambda_n = 0.01$. 
\end{itemize}

\subsubsection{Curvature Loss.} 

While Instant-NSR~\cite{instant-nsr-pl} derives the curvature loss from a discrete Laplacian, we follow a more robust and explicit method as described in PermutoSDF~\cite{rosu2023permutosdf}. For any given point, we randomly perturb it within the tangent plane orthogonal to its normal. 
The curvature loss is then measured by the cosine similarity of normals between the original point and its perturbed counterpart. Note that, we applied the same curvature loss for both Instant-NSR and GSDF.

\subsection{More experiments}
\subsubsection{Reconstruction with 500k iterations.} 

\begin{figure}

  \centering
   \includegraphics[width=1\linewidth]{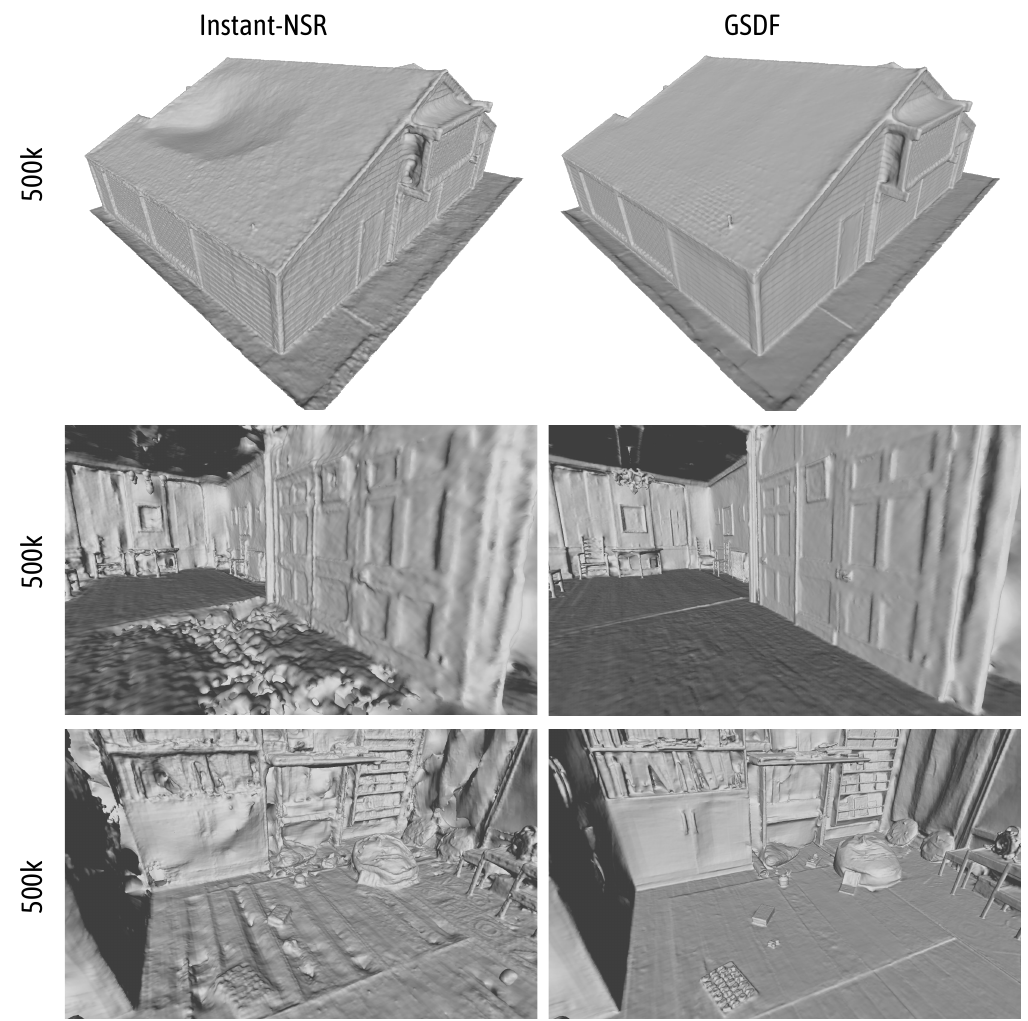}
   
   \caption{\textbf{Reconstruction comparison with $500$k training iterations}. 
   }

   \label{fig:50w}
\end{figure}

As discussed in Sec.~\ref{sec:experiment}, optimizing the SDF is a time-intensive process. While our GS-branch assists in the convergence of the SDF-branch, 
we conjecture that further improvements could be made through additional iterations.
Hence, we train our method for 500$k$ iterations.
As shown in Fig.\ref{fig:50w}, our method achieves superior reconstruction quality compared to Instant-NSR~\cite{instant-nsr-pl} (our SDF-branch) with the same training iterations.

\subsubsection{Rendering comparison with random initialization.} 
\begin{figure}

  \centering
   \includegraphics[width=1\linewidth]{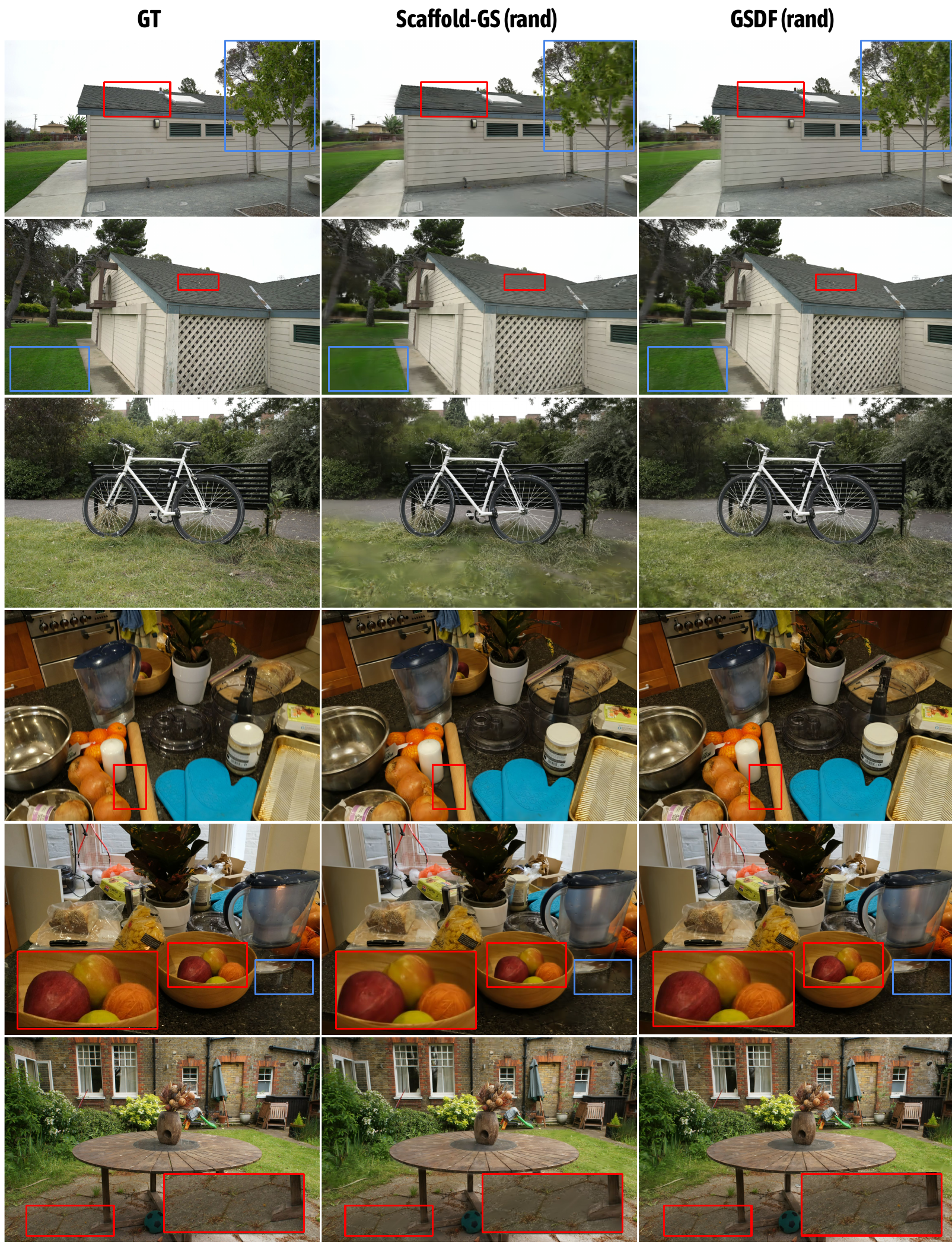}
   
   \caption{\textbf{Rendering Comparison with random initialization (Part 1)}. 
   We compare rendering results between Scaffold-GS~\cite{scaffoldgs} and our method. 
   The highlighted patches indicate that our method is superior in expressing finer details in both geometry and appearance.}

   \label{fig:random1}
\end{figure}

\begin{figure}

  \centering
   \includegraphics[width=1\linewidth]{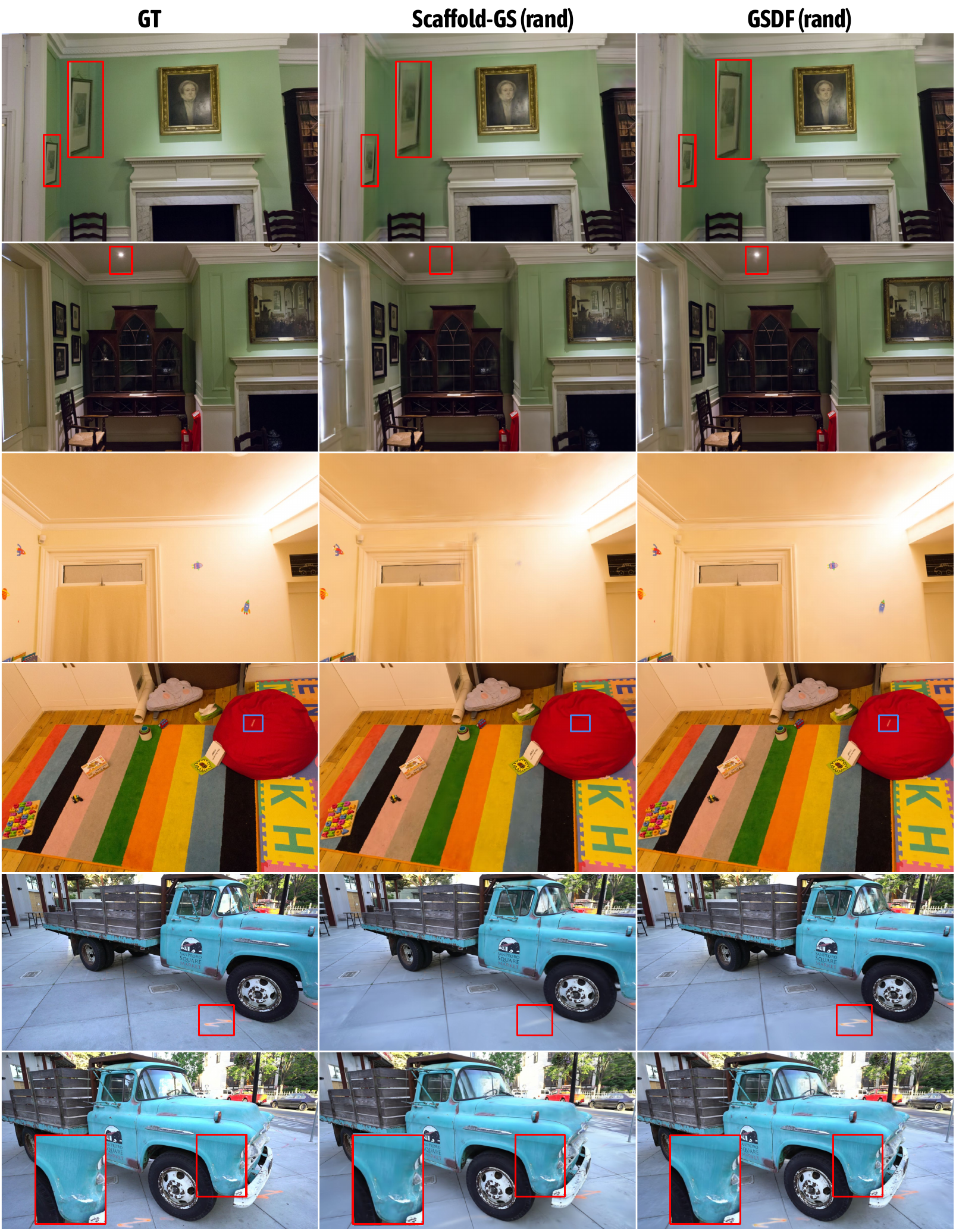}
   
   \caption{\textbf{Rendering Comparison with random initialization (Part 2) - Cont}. 
   }

   \label{fig:random2}
\end{figure}

In Sec.~\ref{sec:dual_branch}, we discussed how the SDF-branch can improve the rendering quality of the GS-branch. This enhancement is particularly noticeable when Gaussian primitives are randomly initialized, as shown in Fig.~\ref{fig:random1} and Fig.~\ref{fig:random2}.

\subsubsection{Per-scene Results.} 
We list the per-scene quality metrics (PSNR, SSIM, LPIPS, and Chamfer Distance) used in our rendering and reconstruction evaluation in Sec.~\ref{sec:experiment} for all considered methods, as shown from Tab.~\ref{tab:a} to Tab.~\ref{tab:c}.

\begin{table}[t!]
\centering
\caption{Quantitative results for DTU scenes \cite{aanaes2016large}. Note that we tried our best, but NeuS~\cite{wang2021neus} failed to reconstruct geometries in scenes $69$ and $105$ with $\frac{7}{8}$ training images.}
\label{tab:a}
\resizebox{\linewidth}{!}{
\begin{tabular}{cccccccccccccccc}
\toprule
\textbf{CD \(\downarrow\)} & 24    & 37    & 40    & 55    & 63    & 65    & 69    & 83    & 97    & 105   & 106   & 110   & 114   & 118   & 122   \\ \midrule
NeuS \cite{wang2021neus}       & \cellcolor{tabthird}0.828 & \cellcolor{tabsecond}0.939 & 1.120 & \cellcolor{tabsecond}0.421 & \cellcolor{tabfirst}1.061 & \cellcolor{tabfirst}0.623 &   -   & \cellcolor{tabsecond}1.474 & \cellcolor{tabfirst}1.180 &  -    & \cellcolor{tabfirst}0.555 & \cellcolor{tabfirst}1.218 & \cellcolor{tabfirst}0.340 & \cellcolor{tabfirst}0.462 & \cellcolor{tabfirst}0.480 \\
Instant-NSR \cite{instant-nsr-pl} & 1.296 & \cellcolor{tabthird}1.107 & 1.284 & 6.582 & 1.459 & \cellcolor{tabsecond}0.710 & 2.361 & \cellcolor{tabthird}1.535 & 1.664 & \cellcolor{tabthird}0.897 & 0.736 & 2.580 & 0.552 & 2.803 & 1.565 \\ \midrule
3D-GS \cite{kerbl20233d}      & 4.877 & 5.720 & 4.738 & 3.133 & 7.119 & 3.713 & 3.486 & 3.830 & 4.294 & 4.323 & 3.218 & 3.418 & 3.648 & 2.445 & 2.550 \\
Scaffold-GS \cite{scaffoldgs} & 7.234 & 6.234 & 6.483 & 7.436 & 8.173 & 4.266 & 5.779 & 5.451 & 6.565 & 6.358 & 5.049 & 5.953 & 6.317 & 5.615 & 2.899 \\
2D-GS \cite{Huang20242DGS}      & \cellcolor{tabsecond}0.662 & 1.108 & \cellcolor{tabsecond}0.707 & \cellcolor{tabthird}0.430 & \cellcolor{tabthird}1.307 & 1.095 & \cellcolor{tabsecond}0.901 & \cellcolor{tabfirst}1.385 & \cellcolor{tabsecond}1.257 & \cellcolor{tabsecond}0.823 & 0.754 & \cellcolor{tabthird}1.419 & \cellcolor{tabthird}0.484 & 0.974 & \cellcolor{tabthird}0.606 \\
SuGaR \cite{guedon2023sugar}      & 1.166 & 1.075 & \cellcolor{tabthird}0.958 & 0.471 & 2.187 & 1.638 & \cellcolor{tabthird}1.022 & 1.812 & 1.505 & 0.930 & \cellcolor{tabthird}0.672 & 2.686 & 0.701 & \cellcolor{tabthird}0.943 & 0.819 \\ \midrule
GSDF        & \cellcolor{tabfirst}0.588 & \cellcolor{tabfirst}0.936 & \cellcolor{tabfirst}0.460 & \cellcolor{tabfirst}0.376 & \cellcolor{tabsecond}1.300 & \cellcolor{tabthird}0.771 & \cellcolor{tabfirst}0.734 & 1.589 & \cellcolor{tabthird}1.287 & \cellcolor{tabfirst}0.760 & \cellcolor{tabsecond}0.588 & \cellcolor{tabsecond}1.222 & \cellcolor{tabsecond}0.379 & \cellcolor{tabsecond}0.518 & \cellcolor{tabsecond}0.513 \\
\bottomrule

\end{tabular}
}

\vspace{1.5em}

\resizebox{\linewidth}{!}{
\begin{tabular}{cccccccccccccccc}
\toprule
\textbf{PSNR \(\uparrow\)}   & 24    & 37    & 40    & 55    & 63    & 65    & 69    & 83    & 97    & 105   & 106   & 110   & 114   & 118   & 122   \\ \midrule
NeuS \cite{wang2021neus}       & 26.96 & 26.08 & 26.86 & 27.89 & 33.76 & 32.90 &   -   & \cellcolor{tabthird}38.13 & 28.95 &   -  & 33.92 & 32.09 & 29.62 & 35.77 & 37.19 \\
Instant-NSR \cite{instant-nsr-pl}& 21.64 & 21.24 & 23.27 & 19.10 & 30.40 & 27.25 & 25.07 & 32.87 & 24.32 & 29.71 & 28.90 & 27.39 & 24.96 & 29.57 & 29.44 \\ \midrule
3D-GS \cite{kerbl20233d}      & 29.98 & 27.33 & 28.87 & 31.64 & \cellcolor{tabfirst}35.68 & 32.76 & 28.75 & 38.05 & 31.78 & \cellcolor{tabsecond}36.63 & 34.66 & 33.08 & 29.94 & 36.83 & 37.78 \\
Scaffold-GS \cite{scaffoldgs} & \cellcolor{tabfirst}30.59 & \cellcolor{tabsecond}27.85 & \cellcolor{tabfirst}29.65 & \cellcolor{tabthird}32.33 & 29.27 & \cellcolor{tabthird}33.22 & \cellcolor{tabthird}29.43 & \cellcolor{tabsecond}38.96 & \cellcolor{tabthird}32.04 & \cellcolor{tabthird}36.60 & \cellcolor{tabsecond}36.38 & \cellcolor{tabfirst}34.05 & \cellcolor{tabfirst}31.07 & 36.67 & 37.73 \\
2D-GS \cite{Huang20242DGS}      & \cellcolor{tabsecond}30.31 & \cellcolor{tabfirst}27.90 & \cellcolor{tabsecond}29.02 & \cellcolor{tabsecond}32.79 & \cellcolor{tabsecond}35.52 & \cellcolor{tabsecond}33.32 & \cellcolor{tabfirst}29.58 & 37.01 & \cellcolor{tabfirst}32.32 & 36.50 & \cellcolor{tabthird}36.25 & \cellcolor{tabthird}33.47 & \cellcolor{tabthird}30.87 & \cellcolor{tabsecond}37.30 & \cellcolor{tabsecond}38.76 \\ \midrule
SuGaR  \cite{guedon2023sugar}     & 28.41 & 26.94 & 28.10 & 32.12 & 34.72 & 32.82 & 28.82 & 38.10 & 31.01 & 35.95 & 36.21 & 32.22 & 30.31 & \cellcolor{tabthird}37.12 & \cellcolor{tabthird}38.66 \\
GSDF        & \cellcolor{tabthird}30.16 & \cellcolor{tabthird}27.65 & \cellcolor{tabthird}28.92 & \cellcolor{tabfirst}32.90 & \cellcolor{tabthird}35.46 & \cellcolor{tabfirst}33.73 & \cellcolor{tabsecond}29.46 & \cellcolor{tabfirst}39.32 & \cellcolor{tabsecond}32.28 & \cellcolor{tabfirst}36.81 & \cellcolor{tabfirst}36.47 & \cellcolor{tabsecond}34.00 & \cellcolor{tabsecond}30.99 & \cellcolor{tabfirst}37.57 & \cellcolor{tabfirst}39.17 \\
\bottomrule

\end{tabular}
}

\vspace{1.5em}

\resizebox{\linewidth}{!}{
\begin{tabular}{cccccccccccccccc}
\toprule
\textbf{SSIM \(\uparrow\)} & 24    & 37    & 40    & 55    & 63    & 65    & 69    & 83    & 97    & 105   & 106   & 110   & 114   & 118   & 122   \\ \midrule
NeuS \cite{wang2021neus}       & 0.780 & 0.816 & 0.749 & 0.884 & 0.950 & 0.962 &   -   & 0.972 & 0.922 &   -   & 0.924 & 0.939 & 0.909 & 0.953 & 0.964 \\
Instant-NSR \cite{instant-nsr-pl} & 0.689 & 0.754 & 0.670 & 0.725 & 0.938 & 0.928 & 0.861 & 0.957 & 0.857 & 0.898 & 0.871 & 0.912 & 0.833 & 0.897 & 0.913 \\ \midrule
3D-GS \cite{kerbl20233d}      & \cellcolor{tabthird}0.915 & 0.897 & 0.887 & 0.951 & \cellcolor{tabthird}0.964 & 0.959 & 0.933 & \cellcolor{tabthird}0.976 & 0.943 & \cellcolor{tabthird}0.957 & 0.952 & 0.952 & 0.934 & 0.962 & 0.971 \\
Scaffold-GS \cite{scaffoldgs} & 0.914 & \cellcolor{tabsecond}0.906 & \cellcolor{tabthird}0.902 & \cellcolor{tabthird}0.956 & 0.948 & \cellcolor{tabthird}0.962 & \cellcolor{tabthird}0.936 & \cellcolor{tabsecond}0.976 & \cellcolor{tabthird}0.945 & \cellcolor{tabsecond}0.957 & \cellcolor{tabthird}0.957 & \cellcolor{tabsecond}0.958 & \cellcolor{tabthird}0.938 & \cellcolor{tabthird}0.965 & 0.972 \\
2D-GS  \cite{Huang20242DGS}     & \cellcolor{tabfirst}0.920 & \cellcolor{tabfirst}0.911 & \cellcolor{tabfirst}0.903 & \cellcolor{tabsecond}0.958 & \cellcolor{tabsecond}0.964 & \cellcolor{tabsecond}0.962 & \cellcolor{tabsecond}0.938 & 0.970 & \cellcolor{tabsecond}0.947 & 0.954 & 0.957 & \cellcolor{tabthird}0.954 & \cellcolor{tabsecond}0.939 & 0.964 & \cellcolor{tabthird}0.973 \\
SuGaR \cite{guedon2023sugar}      & 0.901 & \cellcolor{tabthird}0.903 & 0.889 & 0.955 & 0.956 & 0.955 & 0.932 & 0.974 & 0.939 & 0.952 & \cellcolor{tabsecond}0.958 & 0.947 & 0.935 & \cellcolor{tabsecond}0.965 & \cellcolor{tabsecond}0.973 \\ \midrule
GSDF        & \cellcolor{tabsecond}0.916 & 0.905 & \cellcolor{tabsecond}0.902 & \cellcolor{tabfirst}0.959 & \cellcolor{tabfirst}0.965 & \cellcolor{tabfirst}0.963 & \cellcolor{tabfirst}0.938 & \cellcolor{tabfirst}0.978 & \cellcolor{tabfirst}0.947 & \cellcolor{tabfirst}0.957 & \cellcolor{tabfirst}0.959 & \cellcolor{tabfirst}0.960 & \cellcolor{tabfirst}0.939 & \cellcolor{tabfirst}0.967 & \cellcolor{tabfirst}0.975 \\

\bottomrule

\end{tabular}
}

\vspace{1.5em}

\resizebox{\linewidth}{!}{
\begin{tabular}{cccccccccccccccc}
\toprule
\textbf{LPIPS \(\downarrow\)} & 24 & 37 & 40 & 55 & 63 & 65 & 69 & 83 & 97 & 105 & 106 & 110 & 114 & 118 & 122 \\ \midrule
NeuS \cite{wang2021neus} & 0.780 & 0.816 & 0.749 & 0.884 & 0.950 & 0.962 & - & 0.972 & 0.922 & - & 0.924 & 0.939 & 0.909 & 0.953 & 0.964 \\
Instant-NSR \cite{instant-nsr-pl} & 0.689 & 0.754 & 0.670 & 0.725 & 0.938 & 0.928 & 0.861 & 0.957 & 0.857 & 0.898 & 0.871 & 0.912 & 0.833 & 0.897 & 0.913 \\ \midrule
3D-GS \cite{kerbl20233d} & \cellcolor{tabthird}0.915 & 0.897 & 0.887 & 0.951 & \cellcolor{tabsecond}0.964 & 0.959 & 0.933 & \cellcolor{tabthird}0.976 & 0.943 & \cellcolor{tabsecond}0.957 & 0.952 & 0.952 & 0.934 & 0.962 & 0.971 \\
Scaffold-GS \cite{scaffoldgs} & 0.914 & \cellcolor{tabsecond}0.906 & \cellcolor{tabsecond}0.902 & \cellcolor{tabthird}0.956 & 0.948 & \cellcolor{tabthird}0.962 & \cellcolor{tabthird}0.936 & \cellcolor{tabsecond}0.976 & \cellcolor{tabthird}0.945 & \cellcolor{tabsecond}0.957 & 0.957 & \cellcolor{tabsecond}0.958 & \cellcolor{tabthird}0.938 & \cellcolor{tabthird}0.965 & 0.972 \\
2D-GS \cite{Huang20242DGS} & \cellcolor{tabsecond}0.920 & \cellcolor{tabfirst}0.911 & \cellcolor{tabfirst}0.903 & \cellcolor{tabsecond}0.958 & 0.964 & \cellcolor{tabsecond}0.962 & \cellcolor{tabsecond}0.938 & 0.970 & \cellcolor{tabsecond}0.947 & 0.954 & \cellcolor{tabthird}0.957 & \cellcolor{tabthird}0.954 & \cellcolor{tabsecond}0.939 & 0.964 & \cellcolor{tabsecond}0.973 \\
SuGaR \cite{guedon2023sugar} & 0.901 & 0.903 & 0.889 & 0.955 & 0.956 & 0.955 & 0.932 & 0.974 & 0.939 & 0.952 & \cellcolor{tabsecond}0.958 & 0.947 & 0.935 & \cellcolor{tabsecond}0.965 & \cellcolor{tabthird}0.973 \\ \midrule
GSDF & \cellcolor{tabfirst}0.916 & \cellcolor{tabthird}0.905 & \cellcolor{tabthird}0.902 & \cellcolor{tabfirst}0.959 & \cellcolor{tabfirst}0.965 & \cellcolor{tabfirst}0.963 & \cellcolor{tabfirst}0.938 & \cellcolor{tabfirst}0.978 & \cellcolor{tabfirst}0.947 & \cellcolor{tabfirst}0.957 & \cellcolor{tabfirst}0.959 & \cellcolor{tabfirst}0.960 & \cellcolor{tabfirst}0.939 & \cellcolor{tabfirst}0.967 & \cellcolor{tabfirst}0.975 \\

\bottomrule

\end{tabular}
}
\end{table}

\begin{table}[t!]
\centering
\caption{Quantitative results for Mip-NeRF 360 scenes \cite{barron2022mip}.}
\label{tab:b}
\resizebox{\linewidth}{!}{
\begin{tabular}{c|ccc|ccc|ccc}
\toprule
            & \multicolumn{3}{c|}{bicycle} & \multicolumn{3}{c|}{garden} & \multicolumn{3}{c}{stump} \\
 \textbf{\multirow{-2}{*}{Outdoor}}            & PSNR \(\uparrow\) & SSIM \(\uparrow\) & LPIPS \(\downarrow\)   & PSNR \(\uparrow\) & SSIM \(\uparrow\) & LPIPS \(\downarrow\)  & PSNR \(\uparrow\) & SSIM \(\uparrow\) & LPIPS \(\downarrow\)  \\ \midrule
Instant-NSR \cite{instant-nsr-pl}& 19.68   & 0.449   & 0.615   & 19.74   & 0.422   & 0.627  & 18.77   & 0.486  & 0.592  \\ \midrule
3D-GS  \cite{kerbl20233d}     & \cellcolor{tabthird}24.46   & \cellcolor{tabthird}0.707   & \cellcolor{tabthird}0.313   & \cellcolor{tabsecond}26.63   & \cellcolor{tabsecond}0.819   & \cellcolor{tabsecond}0.175  & \cellcolor{tabsecond}26.31   & \cellcolor{tabsecond}0.758  & \cellcolor{tabthird}0.309  \\
Scaffold-GS \cite{scaffoldgs}& \cellcolor{tabfirst}24.63   & \cellcolor{tabsecond}0.721   & \cellcolor{tabsecond}0.289   & \cellcolor{tabthird}26.59   & \cellcolor{tabthird}0.814   & \cellcolor{tabthird}0.182  & \cellcolor{tabfirst}26.58   & \cellcolor{tabfirst}0.765  & \cellcolor{tabsecond}0.302  \\
Scaffold-GS(rand) & 23.48 & 0.633 & 0.397 & 26.09 & 0.776 & 0.239 & 22.88 & 0.652 & 0.428 \\
2D-GS  \cite{Huang20242DGS}     & 23.96   & 0.683   & 0.358   & 26.07   & 0.803   & 0.211  & 25.69   & 0.746  & 0.349  \\
2D-GS(rand)       & 22.65 & 0.578 & 0.455 & 24.98 & 0.734 & 0.298 & 22.51 & 0.637 & 0.471 \\
SuGaR \cite{guedon2023sugar}      & 23.26   & 0.630   & 0.356   & 25.44   & 0.763   & 0.239  & 25.12   & 0.705  & 0.324  \\ \midrule
GSDF        & \cellcolor{tabsecond}24.61   & \cellcolor{tabfirst}0.724   & \cellcolor{tabfirst}0.261   & \cellcolor{tabfirst}26.91   & \cellcolor{tabfirst}0.824   & \cellcolor{tabfirst}0.158  & \cellcolor{tabthird}26.04   & \cellcolor{tabthird}0.757  & \cellcolor{tabfirst}0.290 \\

GSDF (rand)       & 24.09 & 0.682 & 0.328 & 26.25 & 0.798 & 0.196 & 21.92 & 0.625 & 0.416 \\

\bottomrule
\end{tabular}
}

\vspace{1.5em}

\resizebox{\linewidth}{!}{
\begin{tabular}{c|ccc|ccc|ccc|ccc}
\toprule
            & \multicolumn{3}{c}{room} & \multicolumn{3}{c}{counter} & \multicolumn{3}{c}{kitchen} & \multicolumn{3}{c}{bonsai} \\
\textbf{\multirow{-2}{*}{Indoor} }           & PSNR \(\uparrow\) & SSIM \(\uparrow\) & LPIPS \(\downarrow\)  & PSNR \(\uparrow\) & SSIM \(\uparrow\) & LPIPS \(\downarrow\)   & PSNR \(\uparrow\) & SSIM \(\uparrow\) & LPIPS \(\downarrow\)   & PSNR \(\uparrow\) & SSIM \(\uparrow\) & LPIPS \(\downarrow\)  \\ \midrule
Instant-NSR & 23.18  & 0.769  & 0.439  & 23.00   & 0.723   & 0.459   & 23.87   & 0.685   & 0.400   & 24.84   & 0.753   & 0.430  \\ \midrule
3D-GS \cite{kerbl20233d}      & \cellcolor{tabthird}31.68  & \cellcolor{tabthird}0.925  & \cellcolor{tabthird}0.194  & \cellcolor{tabthird}29.25   & \cellcolor{tabthird}0.914   & \cellcolor{tabthird}0.180   & \cellcolor{tabthird}31.69   & \cellcolor{tabthird}0.933   & \cellcolor{tabthird}0.113   & \cellcolor{tabthird}32.19   & \cellcolor{tabthird}0.945   & \cellcolor{tabthird}0.176  \\
Scaffold-GS \cite{scaffoldgs} & \cellcolor{tabsecond}32.45  & \cellcolor{tabsecond}0.933  & \cellcolor{tabsecond}0.178  & \cellcolor{tabsecond}29.93   & \cellcolor{tabsecond}0.920   & \cellcolor{tabsecond}0.173   & \cellcolor{tabfirst}31.99   & \cellcolor{tabsecond}0.934   & \cellcolor{tabsecond}0.112   & \cellcolor{tabsecond}33.24   & \cellcolor{tabsecond}0.951   & \cellcolor{tabsecond}0.165  \\
Scaffold-GS(rand) & 31.84 & 0.920 & 0.205 & 29.00 & 0.898 & 0.206 & 30.50 & 0.914 & 0.144 & 31.08 & 0.928 & 0.196 \\
2D-GS \cite{Huang20242DGS}      & 30.77  & 0.913  & 0.219  & 28.19   & 0.898   & 0.212   & 30.50   & 0.922   & 0.133   & 31.39   & 0.934   & 0.201  \\
2D-GS(rand)       & 30.01 & 0.895 & 0.256 & 27.78 & 0.874 & 0.248 & 30.10 & 0.914 & 0.147 & 30.56 & 0.922 & 0.222 \\
SuGaR \cite{guedon2023sugar}      & 30.08  & 0.904  & 0.259  & 27.55   & 0.884   & 0.244   & 29.51   & 0.901   & 0.179   & 30.81   & 0.933   & 0.219  \\ \midrule
GSDF        & \cellcolor{tabfirst}32.46  & \cellcolor{tabfirst}0.937  & \cellcolor{tabfirst}0.165  & \cellcolor{tabfirst}30.11   & \cellcolor{tabfirst}0.923   & \cellcolor{tabfirst}0.159   & \cellcolor{tabsecond}31.93   & \cellcolor{tabfirst}0.936   & \cellcolor{tabfirst}0.106   & \cellcolor{tabfirst}33.60   & \cellcolor{tabfirst}0.954   & \cellcolor{tabfirst}0.155 \\
GSDF (rand)       & 32.01 & 0.929 & 0.181 & 29.17 & 0.909 & 0.183 & 31.45 & 0.929 & 0.118 & 31.43 & 0.936 & 0.182 \\
\bottomrule
\end{tabular}
}

\end{table}

\begin{table}[t!]
\centering
\caption{Quantitative results for Tanks\&Temples~\cite{Knapitsch2017} and Deep Blending~\cite{DeepBlending2018} scenes.}
\label{tab:c}
\resizebox{\linewidth}{!}{
\begin{tabular}{c|ccc|ccc|ccc|ccc}
\toprule
Dataset & \multicolumn{6}{c|}{Tanks\&Temple~\cite{Knapitsch2017}} & \multicolumn{6}{c}{Deep Blending~\cite{DeepBlending2018}} \\ \midrule
 & \multicolumn{3}{c|}{Barn} & \multicolumn{3}{c|}{Truck} & \multicolumn{3}{c|}{Dr Johnson} & \multicolumn{3}{c}{Playroom}\\

\multirow{-2}{*}{Method \& Metrics}  & PSNR \(\uparrow\) & SSIM \(\uparrow\) & LPIPS \(\downarrow\) & PSNR \(\uparrow\)  & SSIM \(\uparrow\) & LPIPS \(\downarrow\) & PSNR \(\uparrow\) & SSIM \(\uparrow\) & LPIPS \(\downarrow\) & PSNR \(\uparrow\) & SSIM \(\uparrow\) & LPIPS \(\downarrow\)\\ \midrule

Instant-NSR \cite{instant-nsr-pl} & 20.86  & 0.660  & 0.475  & 17.87   & 0.622  & 0.514  & 24.59                        & 0.791                        & 0.476                     & {23.42} & {0.799} & {0.448} \\ \midrule
3D-GS \cite{kerbl20233d}      & \cellcolor{tabthird}28.21  & \cellcolor{tabthird}0.855  & \cellcolor{tabthird}0.223  & \cellcolor{tabthird}25.24   & \cellcolor{tabthird}0.861  & \cellcolor{tabthird}0.197  & \cellcolor{tabthird}29.04                        & \cellcolor{tabthird}0.897                        & \cellcolor{tabthird}0.248                     & 29.84                        & \cellcolor{tabthird}0.900                        & \cellcolor{tabthird}0.248                        \\
Scaffold-GS \cite{scaffoldgs} & \cellcolor{tabsecond}28.77  & \cellcolor{tabsecond}0.869  & \cellcolor{tabsecond}0.192  & \cellcolor{tabsecond}25.80   & \cellcolor{tabsecond}0.869  & \cellcolor{tabsecond}0.172  & \cellcolor{tabsecond}29.86                        & \cellcolor{tabsecond}0.908                        & \cellcolor{tabsecond}0.236                     & \cellcolor{tabsecond}30.89                        & \cellcolor{tabsecond}0.908                        & \cellcolor{tabsecond}0.239                        \\

Scaffold-GS (rand) & 27.82 & 0.832 & 0.243 & 24.93 & 0.845 & 0.217 & 28.58 & 0.892 & 0.262 & 29.66 & 0.898 & 0.258  \\

2D-GS \cite{Huang20242DGS}      & 26.74  & 0.828  & 0.267  & 24.75   & 0.852  & 0.224  & 28.65                        & 0.895                        & 0.262                     & \cellcolor{tabthird}{29.88} & {0.899} & {0.260} \\
2D-GS (rand)       & 26.48 & 0.810 & 0.288 & 23.84 & 0.824 & 0.262 & 28.13 & 0.889 & 0.276 & 29.40 & 0.895 & 0.273 \\
SuGaR \cite{guedon2023sugar}      & 26.67  & 0.837  & 0.233  & 22.69   & 0.817  & 0.227  & 28.77                        & 0.889                        & 0.271                     & 30.30                        & 0.900                        & 0.258                        \\ \midrule
GSDF        & \cellcolor{tabfirst}28.93  & \cellcolor{tabfirst}0.877  & \cellcolor{tabfirst}0.176  & \cellcolor{tabfirst}25.81   & \cellcolor{tabfirst}0.873  & \cellcolor{tabfirst}0.135  & \cellcolor{tabfirst}{29.87} & \cellcolor{tabfirst}{0.909} & \cellcolor{tabfirst}0.225  & \cellcolor{tabfirst}{30.89} & \cellcolor{tabfirst}{0.909} & \cellcolor{tabfirst}0.222 \\  

GSDF (rand)  & 28.11 & 0.851 & 0.210 & 24.99 & 0.853 & 0.184 & 29.09 & 0.898 & 0.247 & 30.04 & 0.899 & 0.241 \\                      

\bottomrule
\end{tabular}
}
\end{table}

\clearpage

\section*{NeurIPS Paper Checklist}

\begin{enumerate}

\item {\bf Claims}
    \item[] Question: Do the main claims made in the abstract and introduction accurately reflect the paper's contributions and scope?
    \item[] Answer: \answerYes{} 
    \item[] Justification: We introduce GSDF, a dual-branch structure combining the strengths of 3D-GS and SDF, for improving both rendering and reconstruction. Please see our abstract and introduction for further details.

\item {\bf Limitations}
    \item[] Question: Does the paper discuss the limitations of the work performed by the authors?
    \item[] Answer: \answerYes{} 
    \item[] Justification: Our framework still struggles with the dilemma like most SDF-based reconstruction methods. Please see the limitation section (Sec. \ref{sec:limitations}) for further details.

\item {\bf Theory Assumptions and Proofs}
    \item[] Question: For each theoretical result, does the paper provide the full set of assumptions and a complete (and correct) proof?
    \item[] Answer: \answerYes{} 
    \item[] Justification: In the method and experiment sections, we provide a detailed discussion and substantial evidence to demonstrate each result.

    \item {\bf Experimental Result Reproducibility}
    \item[] Question: Does the paper fully disclose all the information needed to reproduce the main experimental results of the paper to the extent that it affects the main claims and/or conclusions of the paper (regardless of whether the code and data are provided or not)?
    \item[] Answer: \answerYes{} 
    \item[] Justification: GSDF is a robust framework that comprises a GS-branch dedicated to rendering and an SDF-branch focusing on learning neural surfaces. By leveraging the mutual guidance between these two branches, our method achieves state-of-the-art performance on both rendering and reconstruction tasks.

\item {\bf Open access to data and code}
    \item[] Question: Does the paper provide open access to the data and code, with sufficient instructions to faithfully reproduce the main experimental results, as described in supplemental material?
    \item[] Answer: \answerNo{} 
    \item[] Justification: We will release the code after acceptance.

\item {\bf Experimental Setting/Details}
    \item[] Question: Does the paper specify all the training and test details (e.g., data splits, hyperparameters, how they were chosen, type of optimizer, etc.) necessary to understand the results?
    \item[] Answer: \answerYes{} 
    \item[] Justification: In the Experimental Setup section (Sec. \ref{sec: exp_setup}) and Implementation details section (Sec. \ref{sec:impl_details}), we provide comprehensive details, including the data splits and hyperparameters used in our experiments.

\item {\bf Experiment Statistical Significance}
    \item[] Question: Does the paper report error bars suitably and correctly defined or other appropriate information about the statistical significance of the experiments?
    \item[] Answer: \answerNo{} 
    \item[] Justification: The experimental result for redering and reconstruction tasks from dense views are very stable, the standard deviation is negligible, so we follow the previous work's practice and not report the error bar.

\item {\bf Experiments Compute Resources}
    \item[] Question: For each experiment, does the paper provide sufficient information on the computer resources (type of compute workers, memory, time of execution) needed to reproduce the experiments?
    \item[] Answer: \answerYes{} 
    \item[] Justification: In Experiment section (Sec. \ref{sec:experiment}), We provide sufficient information on the computer resources, including the type of compute workers GPU and training time.
    
\item {\bf Code Of Ethics}
    \item[] Question: Does the research conducted in the paper conform, in every respect, with the NeurIPS Code of Ethics \url{https://neurips.cc/public/EthicsGuidelines}?
    \item[] Answer: \answerYes{} 
    \item[] Justification: We have carefully reviewed the NeurIPS Code of Ethics and are committed to strictly adhering to its guidelines.

\item {\bf Broader Impacts}
    \item[] Question: Does the paper discuss both potential positive societal impacts and negative societal impacts of the work performed?
    \item[] Answer: \answerYes{} 
    \item[] Justification: Neural rendering and surface reconstruction are crucial tasks with important downstream applications in areas like robotics, physical simulations, and augmented reality. We believe our GSDF system has the potential to advance progress in this domain and support the development of these applications.
    
\item {\bf Safeguards}
    \item[] Question: Does the paper describe safeguards that have been put in place for responsible release of data or models that have a high risk for misuse (e.g., pretrained language models, image generators, or scraped datasets)?
    \item[] Answer: \answerNA{} 
    \item[] Justification: There is no such risks in the rendering and reconstruction tasks.

\item {\bf Licenses for existing assets}
    \item[] Question: Are the creators or original owners of assets (e.g., code, data, models), used in the paper, properly credited and are the license and terms of use explicitly mentioned and properly respected?
    \item[] Answer: \answerYes{} 
    \item[] Justification: In the Experimental Setup section (Sec. \ref{sec: exp_setup}), we cite each used code package and public datasets.

\item {\bf New Assets}
    \item[] Question: Are new assets introduced in the paper well documented and is the documentation provided alongside the assets?
    \item[] Answer: \answerNo{} 
    \item[] Justification: We plan to release the code upon acceptance of the paper.

\item {\bf Crowdsourcing and Research with Human Subjects}
    \item[] Question: For crowdsourcing experiments and research with human subjects, does the paper include the full text of instructions given to participants and screenshots, if applicable, as well as details about compensation (if any)? 
    \item[] Answer: \answerNA{} 
    \item[] Justification: This paper does not involve any crowdsourcing or research with human subjects.

\item {\bf Institutional Review Board (IRB) Approvals or Equivalent for Research with Human Subjects}
    \item[] Question: Does the paper describe potential risks incurred by study participants, whether such risks were disclosed to the subjects, and whether Institutional Review Board (IRB) approvals (or an equivalent approval/review based on the requirements of your country or institution) were obtained?
    \item[] Answer: \answerNA{} 
    \item[] Justification: This paper does not involve any crowdsourcing or research with human subjects.

\end{enumerate}

\end{document}